\documentclass[lettersize,journal]{IEEEtran}
\usepackage{amsmath,amsfonts}
\usepackage{algorithmic}
\usepackage{algorithm}
\usepackage{array}
\usepackage[caption=false,font=normalsize,labelfont=sf,textfont=sf]{subfig}
\usepackage{textcomp}
\usepackage{stfloats}
\usepackage{url}
\usepackage{verbatim}
\usepackage{graphicx}
\usepackage{cite}
\usepackage{multirow}
\usepackage{ulem}
\usepackage{color}
\usepackage{makecell}

\begin{document}

\title{Short-term prediction of construction waste transport activities using AI-Truck}

\author{Meng Xu$^1$ and Ke Han$^{2,*}$
\thanks{This work is supported by the National Natural Science Foundation of China (no. 72071163) and the Natural Science Foundation of Sichuan Province, China (no. 2022NSFSC0474).}
\thanks{$^{1}$M. Xu is with the School of Transportation and Logistics at Southwest
 Jiaotong University, Chengdu, China. $^{2}$K. Han is with the School of Transportation and Logistics at Southwest Jiaotong University, Chengdu, China.}
\thanks{$^*$ Corresponding author, email: kehan@swjtu.edu.cn}
}



\maketitle

\begin{abstract}

Construction waste hauling trucks (or `slag trucks') are among the most commonly seen heavy-duty diesel vehicles in urban streets, which not only produce significant carbon, NO$_{\textbf{x}}$ and PM$_{\textbf{2.5}}$ emissions but are also a major source of on-road and on-site fugitive dust. Slag trucks are subject to a series of spatial and temporal access restrictions by local traffic and environmental policies. This paper addresses the practical problem of predicting levels of slag truck activity at a city scale during heavy pollution episodes, such that environmental law enforcement units can take timely and proactive measures against localized truck aggregation. A deep ensemble learning framework (coined AI-Truck) is designed, which employs a soft vote integrator that utilizes Bi-LSTM, TCN, STGCN, and PDFormer as base classifiers. AI-Truck employs a combination of downsampling and weighted loss is employed to address sample imbalance, and utilizes truck trajectories to extract more accurate and effective geographic features. The framework was deployed for truck activity prediction at a resolution of 1km$\times$1km$\times$0.5h, in a 255 km$^{\textbf{2}}$ area in Chengdu, China. As a classifier, AI-Truck achieves a macro F1 of 0.747 in predicting levels of slag truck activity for 0.5-h prediction time length, and enables personnel to spot high-activity locations 1.5 hrs ahead with over 80\% accuracy.
\end{abstract}

\begin{IEEEkeywords}
Slag truck control, Traffic flow category prediction, Deep learning, Ensemble learning, Big data.
\end{IEEEkeywords}

\section{Introduction}
\IEEEPARstart{A}{ccording} to the World Health Organization (WHO), 99\% of the global population is exposed to air that surpasses the recommended WHO guidelines. Prolonged exposure to poor air quality can lead to severe health problems \cite{ref1, ref2}. Research \cite{ref3} suggests that air pollution has now become the fourth leading cause of death worldwide. Traffic is the primary source of urban air pollution such as NO$_x$ \cite{refnew1}, and contributing to 25\% of PM$_{2.5}$ on a global scale \cite{ref5, ref6}. In China, on-road vehicles contribute 59\% of total NO$_x$ emissions, among which heavy-duty diesel vehicles (HDDVs) account for over 76\% of NO$_x$ and 54\% of PM$_{2.5}$ of all traffic emissions \cite{ref65}.

In many developing countries, construction waste hauling trucks (or {\it slag trucks}) are the most commonly seen HDDVs, in terms of not only quantity (e.g. there are over 16,000 slag trucks in the city of Chengdu), but also distribution (they can enter any part of the city that has construction activities). These slag trucks emit significant NO$_x$ and PM$_{2.5}$, and are notable contributors to air pollution, as substantiated by various research \cite{ref4, refMG2015, refSMMSP2022, refMBS2021}. In addition to tailpipe emissions, slag trucks are also known for generating fugitive dust during the transportation of soil and sand \cite{ref8, ref9, ref10}, which is a significant contributor to atmospheric particulate matter \cite{ref7}. Last but not least, slag truck aggregation usually indicates ongoing earthworks throughout the city, which are also within the purview of environmental management. 

To combat air pollution, local authorities have implemented a range of transportation and environmental policies targeting the use of slag trucks, especially during {\it heavy pollution episodes} (HPEs). Specifically, a series of access restrictions are imposed for several {\it key management areas} (KMAs) in the city during HPEs. From the standpoint of environmental law enforcement, it is crucial to predict locations with high truck concentration, such that personnel can be dispatched in a timely fashion to collect evidence and administer on-site intervention, thereby enabling proactive measures against environmental deterioration associated with the use of slag trucks. 

Driven by this practical challenge, this paper focuses on short-term prediction of the levels of slag truck activity on a city scale during heavy pollution episodes. Specifically, we aim to predict the concentration of {\it stay points} \cite{PTG2016}, which are associated with on-site operations of slag trucks (e.g., loading/unloading). We propose a deep ensemble learning framework, {\bf AI-Truck}, to assist environmental law enforcement, which employs a soft vote integrator that utilizes Bi-LSTM, TCN, STGCN, and PDFormer as the base models. Our work demonstrates its scientific and practical values in the following aspects: 
\begin{enumerate}

\item This work uses bagging with a soft voting strategy to synthesize information from multiple independent base models (deep neural networks), which renders more stable and reliable prediction results than a single model.

\item Unlike conventional traffic prediction that focuses on link flows or speeds, this work predicts vehicle concentration in a two-dimensional space, which has a unique challenge of data imbalance due to the sparse (imbalanced) spatial distribution of truck activities. We address this issue by proposing a combination of downsampling and weighted loss.

\item Diverging from previous works that assume correlated flows in all neighboring grids, our method utilizes slag truck trajectories to identify correlations between neighboring grids, which results in a more accurate and effective geographic features.

\item In a real-world scenario in Chengdu (China) during heavy pollution episodes in August 2022, the predictions achieve a macro F1 of 0.747 in predicting levels of slag truck activity for 0.5-h prediction time length. A real-world case study suggest that AI-Truck can assist law enforcement units identify high-activity locations 1.5 hrs ahead with over 80\% accuracy. 

\end{enumerate}
 
\section{Related work}
\subsection{Slag Trucks}
\noindent Slag trucks, a specialized type of heavy-duty diesel vehicles used for transporting solid waste such as construction debris, nowadays have gained considerable attention from scholars, leading to extensive research from different perspectives and scales due to their crucial impact on the traffic order and ecological environment of the areas they traverse \cite{refWYYBL2022}. Lu \textit{et al.} \cite{refL2019} employed data mining techniques to identify illegal solid waste dumping incidents, Lu \textit{et al.}  \cite{refLYL2022} investigated the patterns and influencing factors of loading process on slag trucks, Yu \textit{et al.}  \cite{refLK2024} classified the frequent spots of slag trucks based on machine learning and Wei \textit{et al.} \cite{refWYYBL2022} studied the load and carbon emission characteristics of slag trucks. In this paper, we propose a novel deep ensemble learning framework for predicting the levels of slag truck activity during heavy pollution episodes, which is an unexplored research gap.

\subsection{Traffic Prediction}
\noindent Traffic prediction is a challenging task due to the complex and dynamic spatio-temporal dependencies of traffic patterns, i.e., the spatial dependencies over the graph structure and the temporal dependencies along the time. To capture these dependencies, neural networks (NNs), especially spatio-temporal graph neural networks (STGNNs), have recently become a favorite due to their remarkable learning capability \cite{ref12,ref13,ref14}. Yao \textit{et al.} \cite{ref16} proposed a novel approach for taxi demand prediction based on a multi-perspective (i.e., temporal view, spatial view and semantic view) spatio-temporal neural network, Yu \textit{et al.} \cite{ref19} designed a purely convolutional spatio-temporal graph neural network for traffic flow prediction, which could achieve faster training with fewer parameters and Fang \textit{et al.} \cite{ref22} proposed spatial-temporal graph ordinary differential equation networks for traffic flow prediction, which better captures the spatio-temporal dependencies through tensor-based ordinary differential equations. However, NNs often suffer from instability issues\cite{refYLSDY2017, refXJ2020}. Therefore, we employ bootstrap aggregation (bagging), which is a powerful ensemble technique that combines multiple weak models (base models) to create a strong model (ensemble model), enhancing the stability and accuracy of classification and regression algorithms \cite{ref28, ref29}.

According to previous studies, traffic prediction tasks can be generally divided into two types: traffic flow prediction and traffic speed prediction. Choi \textit{et al.} \cite{ref23}, Li \textit{et al.} \cite{ref24} and Guo \textit{et al.} \cite{ref25} have conducted a lot of explorations in traffic flow prediction, while traffic speed prediction has been explored by Yu \textit{et al.} \cite{ref19}, Liu \textit{et al.} \cite{refQBY2018} and Jia \textit{et al.} \cite{refYJY2016}. However, we note that there is a lack of prediction of the vehicle concentration, which is exactly what we do in the paper. 
\IEEEpubidadjcol

\section{Data and problem description}\label{Data and problem description}
\subsection{Data Description}
\noindent Approximately 14,000 slag trucks are tracked with GPS sensors transmitting trajectory data every 30 seconds. Given the background of our work, which is related to environment law enforcement, we focus on the heavy pollution episode from August 3, 2022, 0:00 to August 28, 2023, 23:00. 

\begin{figure}[htb]
\centering
\includegraphics[width=3.0in]{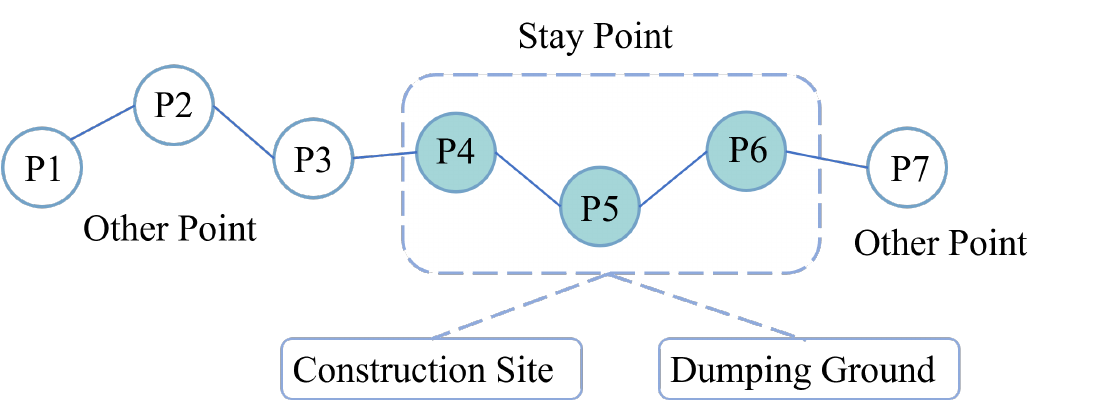}
\caption{Regional categories of stay points.}
\label{fig1}
\end{figure}

As explained in the introduction, we are interested in predicting the concentrations of stay points, because these locations are often construction sites or dumping grounds (as illustrated in Figure \ref{fig1}), which are associated with construction waste transport activities. Consequently, it would be most effective for personnel to collect evidence and administer intervention at these locations. Following \cite{PTG2016}, we define a stay point as the collection of trajectory points $(t_i,\,lat_i,\, lon_i), \,1\leq i\leq n$ of a given truck such that 
$$
\text{distance}\big((\text{lat}_1,\, \text{lon}_1),\,(\text{lat}_j,\,\text{lon}_j)\big)\leq \delta\quad\forall 1\leq j\leq n 
$$
and 
$$
t_n-t_1\geq \theta
$$
In this paper, $\delta=200$m and $\theta=10$min. In other words, a stay point is registered with the truck if the trajectories within a 10-min period are no more than 200m from one another.
 
\begin{figure}[htb]
	\centering
	\includegraphics[width=0.45\textwidth]{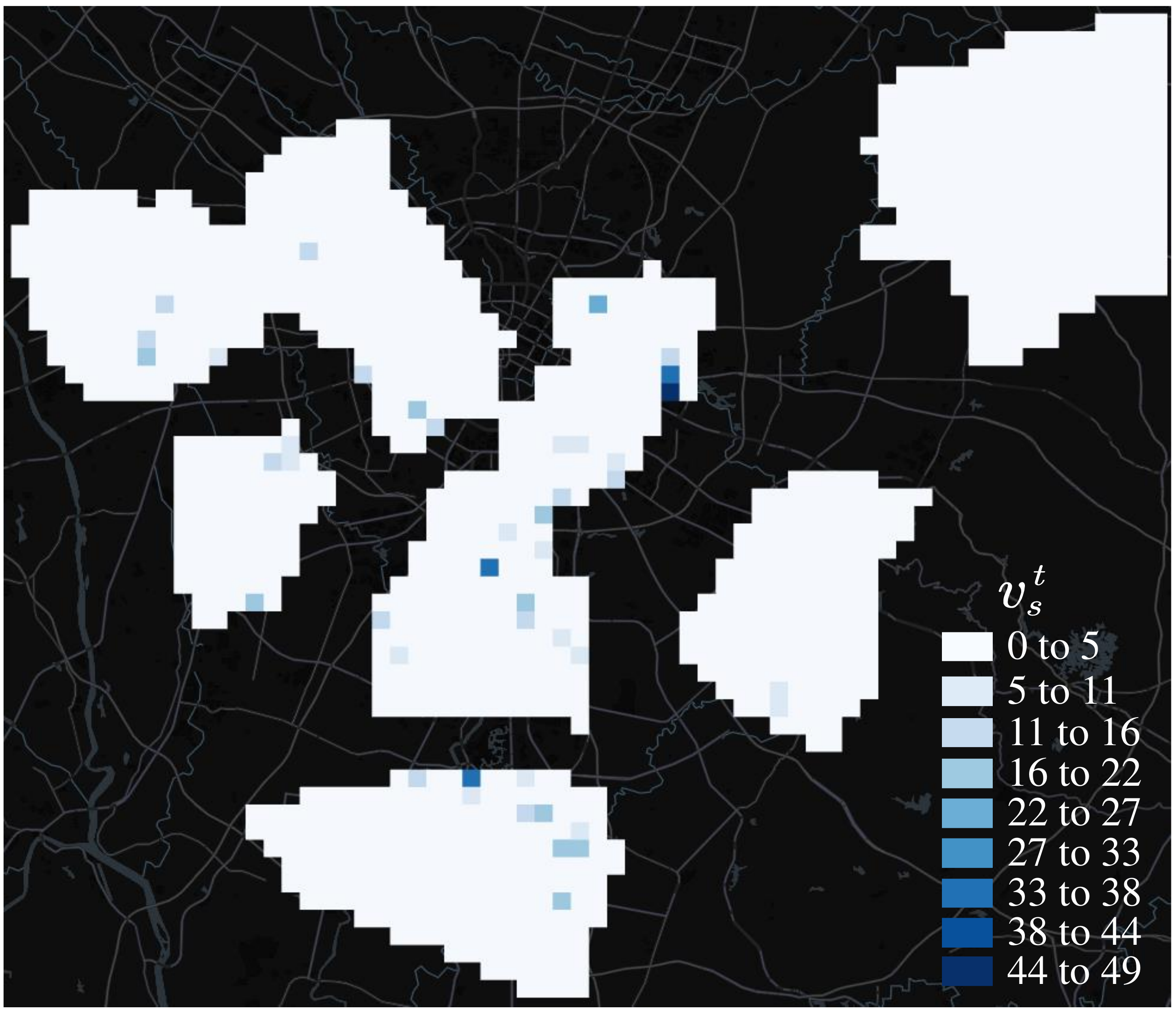}
	\caption{Spatial distribution of $v_s^{t}$ in Chengdu's KMAs.}
	\label{fig2}
\end{figure}

The target region, namely key management areas (KMAs) in Chengdu, is partitioned into 1,199 grids $s\in S$, each with the dimension 1km$\times$1km, as shown in Figure \ref{fig2}. In addition, time is discretized into 0.5 hour intervals $t\in T$. The number of slag truck stay points within each grid $s$ at time slot $t$ is denoted $v_s^{t}$.

Figure \ref{fig2} depicts the spatial distribution of $v_s^{t}$ over the KMAs (colored areas), with darker shades indicating higher values, which clearly shows an imbalanced spatial distribution as most grids have low $v_s^{t}$ values. Such imbalance is due to the fact that slag trucks tend to serve specific locations, such as construction sites and dumping ground. Unfortunately, this imbalance will affect the model's ability to accurately predict areas with high concentration of slag trucks, which are minority situations. Therefore, we adopt a sampling method in our data processing phase to mitigate the imbalanced spatial distribution of slag trucks.

\begin{figure}[htb]
	\centering
	\includegraphics[width=3.0in]{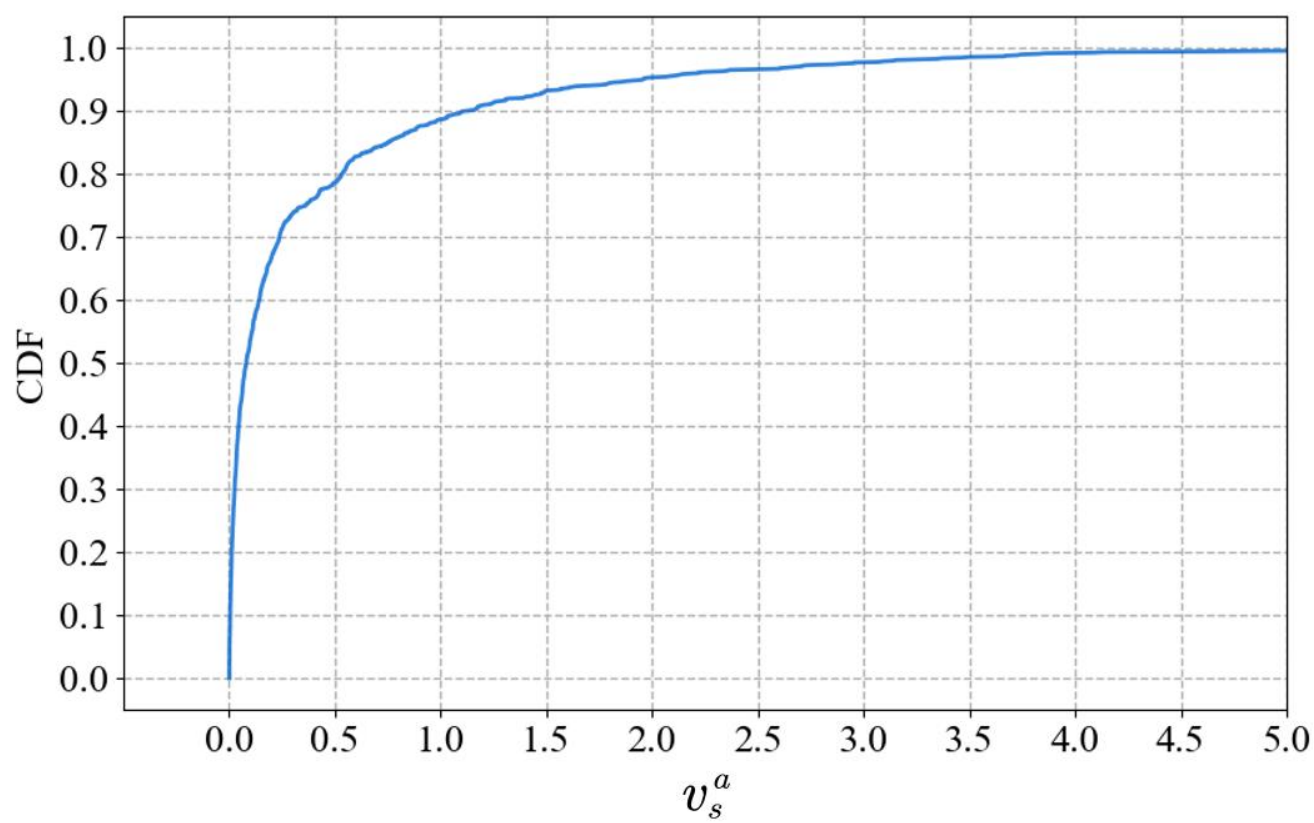}
	\caption{Cumulative distribution function of $v_s^a$ during heavy pollution episodes.}
	\label{figcdf_grid}
\end{figure}

\begin{figure}[htb]
	\centering
	\includegraphics[width=0.45\textwidth]{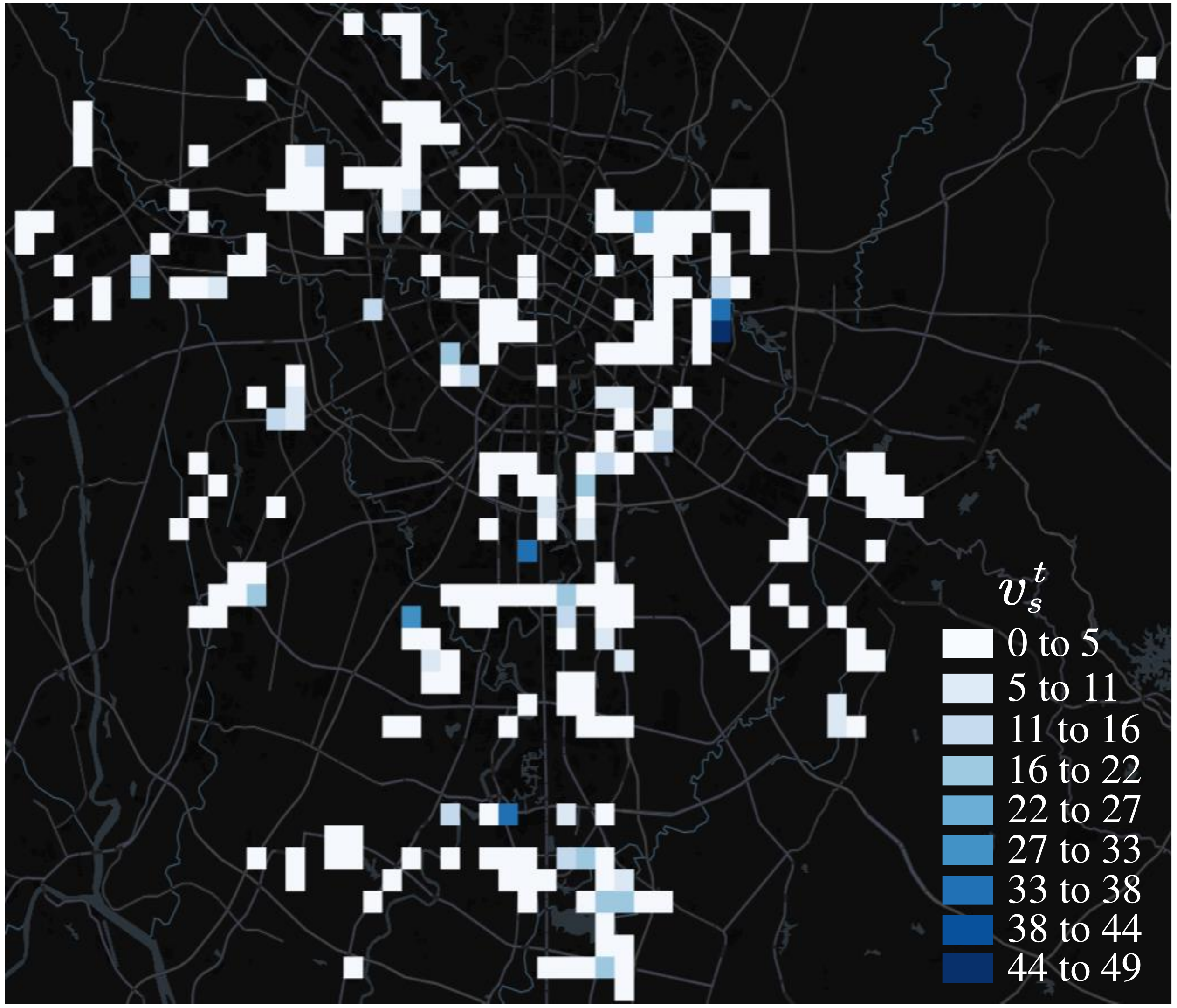}
	\caption{Spatial distribution of $v_s^{t}$ after down-sampling in Chengdu's KMAs.}
	\label{fig3}
\end{figure}

Upsampling and downsampling are the two most common sampling methods, but upsampling is not feasible due to the complex and dynamic spatio-temporal dependencies of slag trucks. In addition, the model predictions rely on features at successive time slots, so downsampling in the temporal dimension is also not feasible. For these reason, the best choice is to downsample in the spatial dimension. Therefore, we employ the cumulative distribution function (CDF) to identify grids with relatively high concentration of slag trucks, and only use them as our target region. Specifically, we calculate the statistical average of $v_s^{t}$ for each grid $s$ during heavy pollution episodes (HEPs), i.e., $v_s^{a}$, and introduce an empirical distribution of it. As shown in Figure \ref{figcdf_grid}, 20\% of the grids consistently lack stay points ($v_s^{a}=0$) during HEPs, indicating that this part of the grid has low predictive value and can be dropped directly. While for those grids with $v_s^{a}>0$, we only select the top 25\% of them as our target region (corresponding to $v_s^a>0.36$). As a result, we ultimately select 255 grids as our study area, as shown in Figure \ref{fig3}, where darker shades represent higher values of $v_s^{t}$. Figure \ref{fig3} demonstrates that the issue of imbalanced spatial distribution has been mitigated following the downsampling.

\subsection{Problem Definition}
\noindent Given $k$ historical states (features) $\left( f^{t-k+1},...,f^{t} \right)$ about slag trucks' activities, this work aims to predict the levels of slag truck activity $c^{\tau}$ where 
$$
c^{\tau}=\big(c_s^{\tau}\big)_{s\in S}\quad \tau >t
$$

Here, $c_s^{\tau}$ denotes the level of slag truck activity, which are categorical values based on $v_s^{\tau}$, the number of slag trucks with staying activities in grid $s$ during time $\tau$. To properly define these categorical values, we invoke the empirical distribution of $v_s^{\tau}$ during heavy pollution episodes. 
\begin{figure}[htb]
	\centering
	\includegraphics[width=3.0in]{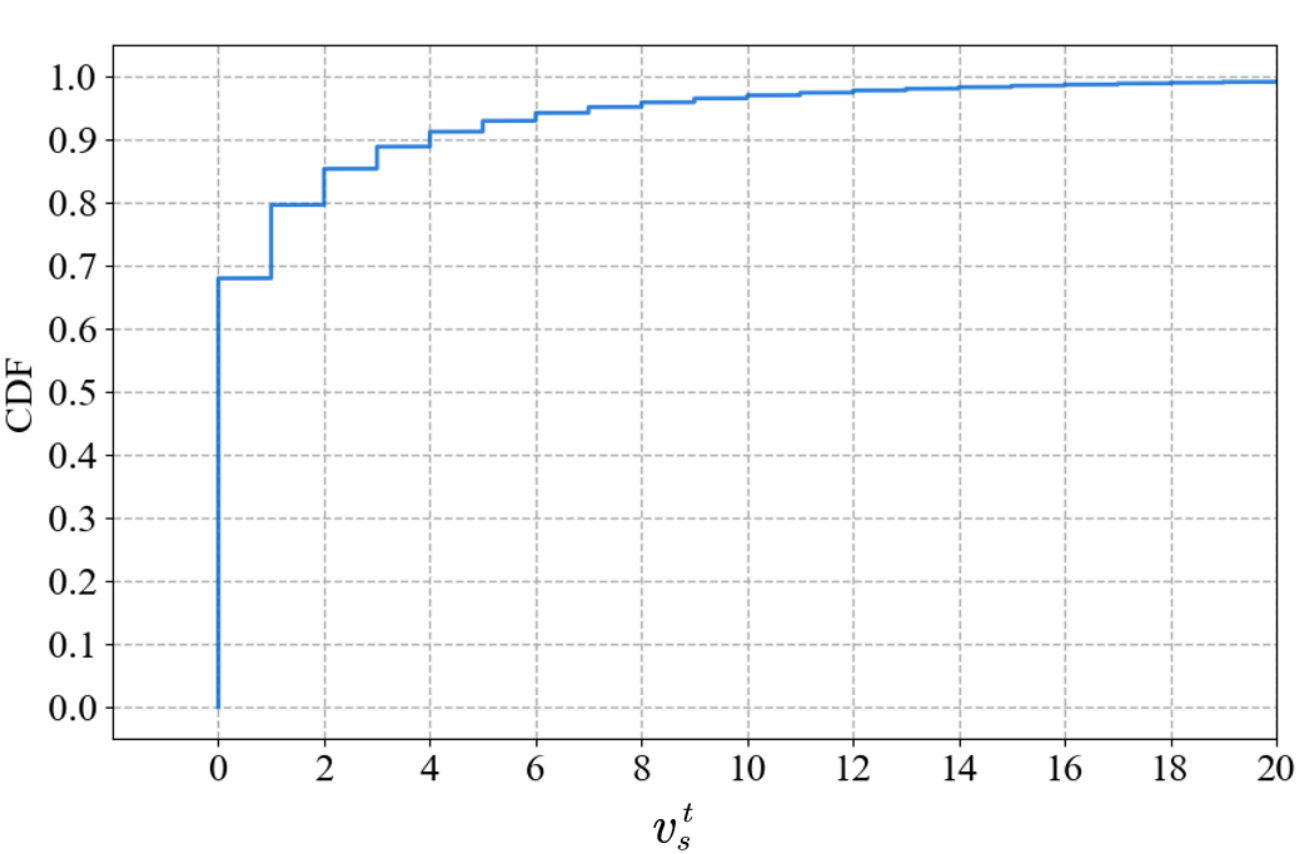}
	\caption{Cumulative distribution function of $v_s^t$ during heavy pollution episodes.}
	\label{figcdf}
\end{figure}
As Figure \ref{figcdf} shows, the majority (68\%) of $v_s^t$ is zero during heavy pollution episodes. For $v_s^t>0$, which take up 32\% of the target, we are only interested in those top 10\% (corresponding to $v_s^t>4$), as resources of environmental law enforcement units are limited. Such a practical consideration leads to the following definition:
\begin{equation}\label{eq2}
c_s^t=
\begin{cases}
0 ~\text{(none)}\quad & \text{if}~v_{s}^{t}=0, 
\\
1 ~\text{(medium)}\quad & \text{if}~v_{s}^{t}\leq 4, 
\\
2~ \text{(high)}\quad & \text{if}~v_{s}^{t}> 4.
\end{cases}
\end{equation}

\section{SPATIAL AND TEMPORAL FEATURES}
\subsection{Spatial Dependencies}
\noindent By effectively capturing spatial dependencies, prediction accuracy could be significantly improved \cite{ref38}. Specifically, traffic patterns demonstrate geographic and semantic dependencies in space, so we construct spatial features from both geographic and semantic perspectives.

\subsubsection{Geographic Dependency}
\begin{figure}[htb]
\centering
\includegraphics[width=.25\textwidth]{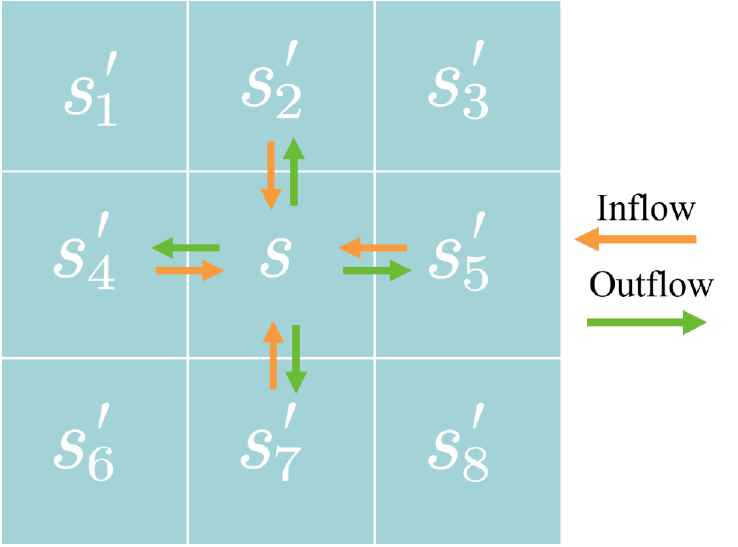}
\caption{Geographic dependency between neighboring grids.}
\label{fig5}
\end{figure}

In grid-based traffic maps, there is a robust dependency between the grid and its neighboring grids \cite{ref39, ref40}. Specifically, the flow in grid $s$ is influenced by both the inflow from and outflow to its neighboring grids $s\prime_m$, as shown in Figure \ref{fig5}. Therefore, it can be inferred that the levels of slag truck activity in grid $s$ is influenced by the levels of slag truck activity in its neighboring grids. 

A common assumption in previous studies is that the grid $s$ has geographic dependency on all its neighboring grids $s\prime_m$ \cite{ref15, ref26, ref39, ref41}. However, this assumption is incorrect, as the levels of slag truck activity on grid $s$ depend on many other factors, such as road network, leading to the slag truck only flow to a few specific neighboring grids instead of all of them. Therefore, we employ consecutive trajectory points to determine whether there is a geographic dependency between the neighboring grids. Specifically, if a slag truck travels from one grid to another in two successive trajectory points, we can infer an origin-destination (OD) relationship (geographic dependency) between two neighboring grids, and the adjacency matrix $A$ based on geographic dependency can be constructed as follows:
\begin{equation}\label{eq3}
A\left( i, j \right)=
\begin{cases}
1,\quad& \text{if two neighboring grids have OD relationship}
\\
0,\quad &\text{otherwise}
\end{cases}
\end{equation}

\subsubsection{Semantic Dependency}
\begin{figure}[htb]
\centering
\includegraphics[width=0.5\textwidth]{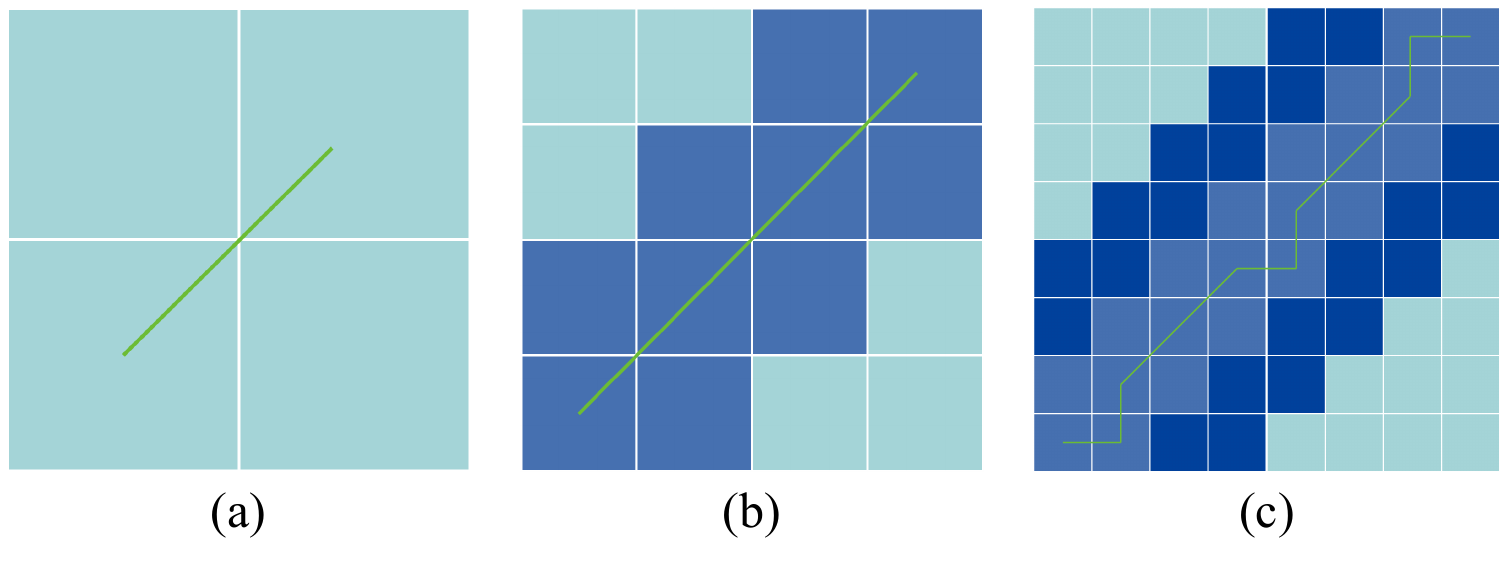}
\caption{The schema of FastDTW. (a) Time granularity is 4/1. (b) Time granularity is 4/1. (c) Time granularity is 1/1.}
\label{fig6}
\end{figure}

Although all grids in the study area have relatively high concentration of slag trucks, their operation patterns are vary from each other due to the factors such as location type, site size and stage of construction. Therefore, these grids demonstrate different degrees of semantic dependency with each other. To explore their semantic dependency, we adopt fast dynamic time warping (FastDTW), which is an approximation algorithm for dynamic time warping (DTW) but has linear time and space complexity \cite{ref42}. FastDTW is a popular method for measuring the similarity between time series data, as it can handle different lengths and alignments of the data. Its core idea and calculation process are shown in \eqref{eq4} and Figure \ref{fig6}, respectively. Finally, we use a semantic matrix $D\left( i,j \right) $ to express the semantic dependency between grids. 
\begin{equation}\label{eq4}
\begin{aligned}
D\left( i,j \right) =&\text{Dist}\left( i,j \right)+\min \\&\big\{ D\left( i-1,j \right) ,D\left( i,j-1 \right) ,D\left( i-1,j-1 \right) \big\} 
\end{aligned}
\end{equation}
\noindent where $\text{Dist}\left( i,j \right) $ represents the first-order relationship between sequences $i$ and $j$.

\subsection{Temporal Dependencies}
\noindent Temporal proximity and periodicity are two crucial components of the temporal dependencies in traffic patterns \cite{ref45}, both of which play an important role in traffic prediction, particularly temporal periodicity \cite{ref46}.

\subsubsection{Temporal Proximity}
\begin{figure}[htb]
\centering
\includegraphics[width=.5\textwidth]{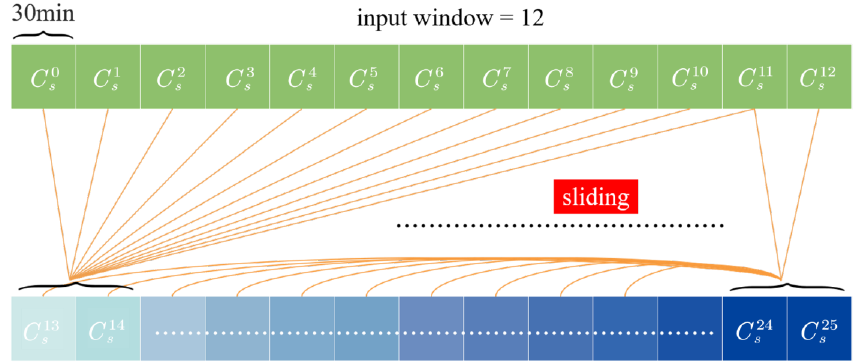}
\caption{Sliding window schematic.}
\label{fig8}
\end{figure}

To capture the temporal proximity in the levels of slag truck activity, we apply a sliding window technique \cite{ref47} with a window size of $k=12$, as shown in Figure \ref{fig8}.

\subsubsection{Temporal Periodicity}
\begin{figure*}[htb]
\centering
\includegraphics[width=.8\textwidth]{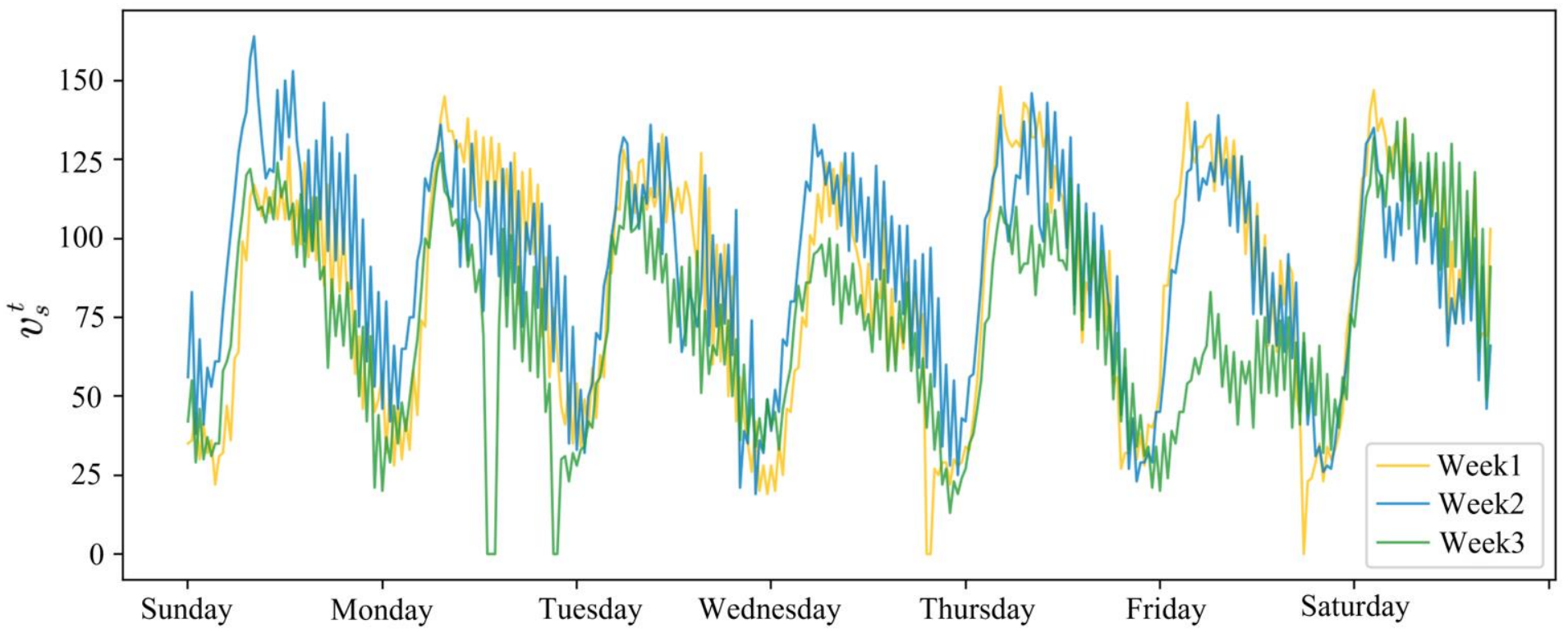}
\caption{Temporal periodicity of $v_s^t$.}
\label{fig9}
\end{figure*}

A model that effectively captures temporal periodicity can predict future traffic conditions more accurately \cite{ref46}. Figure \ref{fig9} illustrates the variation of $v_s^t$ over three consecutive weeks during the heavy pollution episodes, which clearly reveals that $v_s^t$ exhibits the daily and weekly periodicity. Based on it, we can infer that there is a similar pattern in the levels of slag truck activity. Moreover, compared to other traffic flows, the flow of slag trucks has more unique characteristics. Specifically, slag truck flow lacks difference between the weekday and weekend, as these truck are required to work all year round. In addition, slag truck flow exhibits a high degree of instability, as these trucks are easily affected by external factors, such as policy, which poses a challenge to our work.

To capture the daily and weekly periodicity, we extracted the information $h$ and $w$ from the time slots, which indicate the hour of day and the day of week, respectively.

\subsection{Features Summary}
\noindent To summarize, we present the following expression for the spatio-temporal features at a given grid $s$ and time $t$:
\begin{align}
f_{p}^{t}&=\left( A,D \right)  \\
f_{m}^{t}&=\left( h^{t},w^{t} \right) \\
f_{s}^{t}&=\left( c_s^t, f_{m}^{t},f_{p}^{t} \right)
\end{align}
\noindent where $f_{p}^{t}$, $f_{m}^{t}$, and $f_{s}^{t}$ denote the spatial, temporal, and spatio-temporal features, respectively.

\section{Deep Ensemble Learning Frameworks}
\begin{figure*}[htb]
\centering
\includegraphics[width=.9\textwidth]{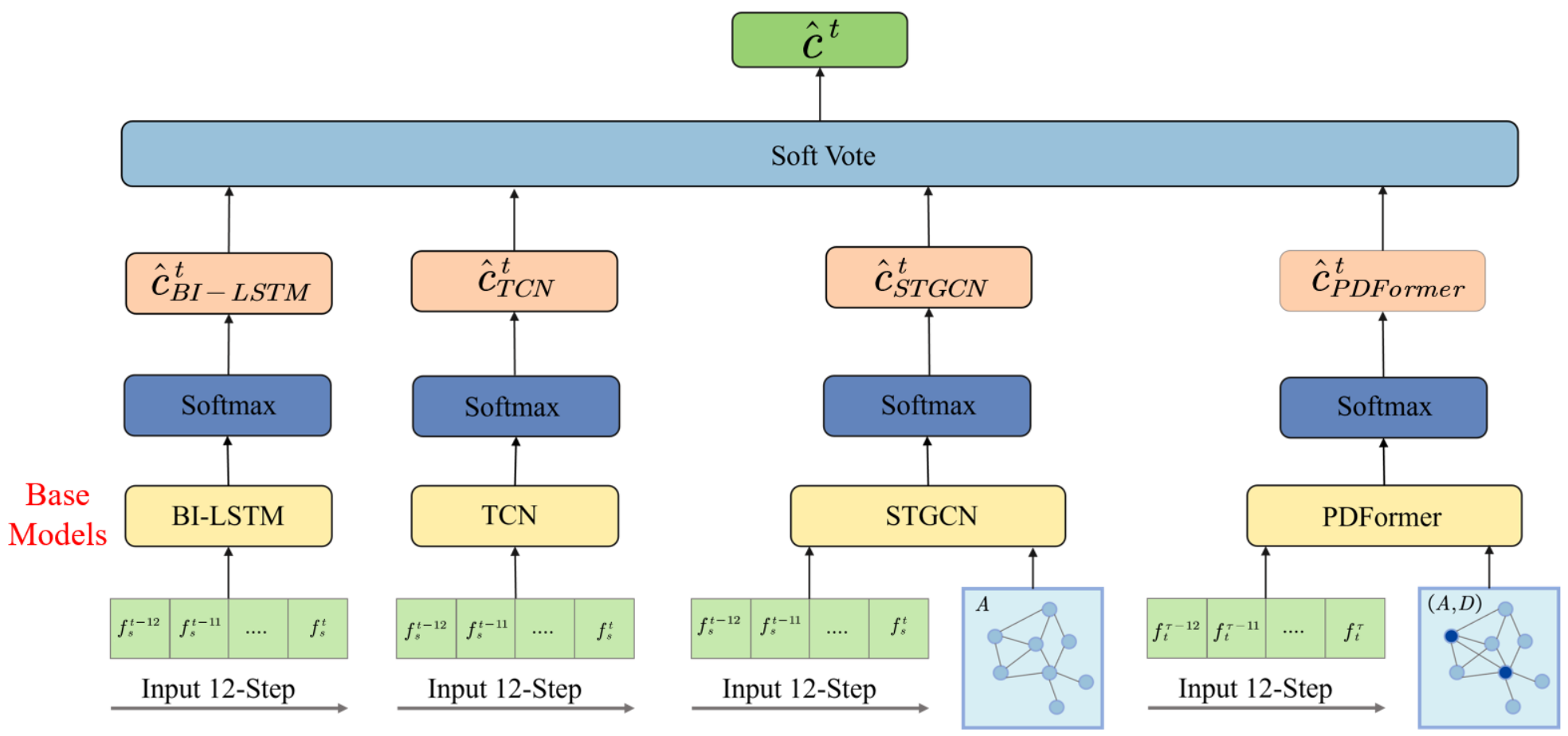}
\caption{The framework of AI-Truck. AI-Truck employs soft voting as an integrator and Bi-LSTM, TCN, STGCN and PDFormer as base models to predict the levels of slag truck activity. Among these base models, Bi-LSTM and TCN only exploit temporal features, while STGCN adopts both temporal and geographical feature, and PDFormer utilizes all the spatio-temporal features mentioned in this paper.}
\label{fig10}
\end{figure*}

\noindent We employ AI-truck (as shown in Figure \ref{fig10}), a deep ensemble learning framework, to predict the levels of slag truck activity. Specifically, we adopt a bagging approach that integrates the outputs of multiple base models to improve the accuracy and stability of AI-truck predictions. The key to bagging is the diversity of the base models, which enables them to compensate for each other’s errors and achieve a more accurate output \cite{ref30}. Therefore, we select bi-directional long short-term memory (Bi-LSTM) \cite{refMK1997}, temporal convolutional network (TCN) \cite{refCMRAG2017}, spatio-temporal graph convolutional network (STGCN) \cite{ref19}, and propagation delay-aware dynamic long-range transformer (PDFormer) \cite{ref26} as the base models for AI-Truck. These models adopt different types of neural networks and have different theoretical foundations, which fits well with the bagging's requirements for base models. Moreover, these models are the NNS, a type of deep learning model, which can effectively capture spatio-temporal dependencies of traffic patterns. 

Soft voting is one of the bagging strategy that could improves the accuracy and stability of the ensemble model by softly aggregating the outputs of multiple base models, i.e., calculating their weighted average. Therefore, we apply a soft vote integrator to aggregate the output of base models ($\hat{c}_{\text{model}}$), and the aggregation principle is as follows:
\begin{equation}\label{eq8}
\begin{aligned}
\hat{c}^{t}&=w_{\text{Bi-LSTM}}\times \hat{c}_{\text{Bi-LSTM}}^{t}+w_{\text{TCN}}\times \hat{c}_{\text{TCN}}^{t}
\\
&+w_{\text{STGCN}}\times \hat{c}_{\text{STGCN}}^{t}+w_{\text{PDFormer}}\times \hat{c}_{\text{PDFormer}}^{t}
\end{aligned}
\end{equation}

\noindent where $w_{\text{model}}$ is the weight of the base model. 

\section{Experiment}
\subsection{Experimental Setup}
\subsubsection{Dataset Descriptions}
We carry out our experiments on a real-world spatio-temporal datasets in Chengdu, China, as mentioned in section \ref{Data and problem description}.

\subsubsection{Baseline Methods}
We compare the performances of AI-Truck with five baseline methods as follows:
\begin{itemize}
\item \textbf{Bi-LSTM}: Bi-directional long short-term memory designs two LSTMs that captures temporal patterns of time series data from both positive direction (forward states) and negative direction (backward states) \cite{refMK1997}.

\item \textbf{TCN}: Temporal convolutional network utilizes a hierarchy of temporal convolutional filters to capture long-range temporal patterns in time series data \cite{refCMRAG2017}.

\item \textbf{STGCN}: Spatio-temporal graph convolutional network is a purely convolutional architecture that uses temporal convolutional network to capture temporal dependencies and graph convolutional network to capture spatial dependencies, thus enabling faster training with fewer parameters \cite{ref19}.

\item \textbf{PDFormer}: Propagation delay-aware dynamic long-range transformer is a one of the latest state-of-the-art models for traffic prediction, which employ a transformer-based architecture to capture delayed propagation in traffic patterns.
\end{itemize}

\subsubsection{Evaluation Strategy and Parameters}
Although the imbalanced spatial distribution has been mitigated, it still remains, with $c_{s}^{t}=0$ accounts for 68\% of all samples. To further address the problem of imbalanced spatial distribution, we adopt a weighted loss approach. Specifically, we improve the model's predictive accuracy of minority samples by assigning higher weights to them in the loss function, which is defined as follows:
\begin{equation}\label{eq9}
	\begin{aligned}
		\text{loss}=-\sum_{i=0,1,2}{w_{i}\times \log \left( p_i \right)}
	\end{aligned}
\end{equation}
\noindent where $w_{i}$ and $p_i$ represent the weight and probability for the level of slag truck activity $c_{s}^{t}=i$.

To evaluate the model's performance and ensure result reliability, we conduct ten experiments using different random seeds, and report the mean and standard deviation of precision, recall, F1 score and macro F1 across the 10 trials as model's final results. For each experiment, we split the data into training and testing sets following an 8:2 ratio, and utilize data from the preceding 6 hours (12 time slots) to predict the levels of slag truck activity in next time step (13th time slot).

AI-Truck, Bi-LSTM, TCN, STGCN and PDFormer are implemented with PyTorch 1.13.1 and trained on NVIDIA GeForce 4060 GPU with the Adam optimizer. Additionally, we perform an extensive search for parameters, AI-Truck is optimized when batch size is $16$, $\left( w_0,w_1,w_2 \right) =\left( 0.7,1.2,1.1 \right)$ and  $\left( w_{\text{Bi-LSTM}},w_{\text{TCN}},w_{\text{STGCN}},w_{\text{PDFormer}} \right) =\left( 1.1,1.1,0.5,1.3 \right)$. 

\subsection{Model Comparison}
\begin{figure}[htb]
	\centering
	\includegraphics[width=.35\textwidth]{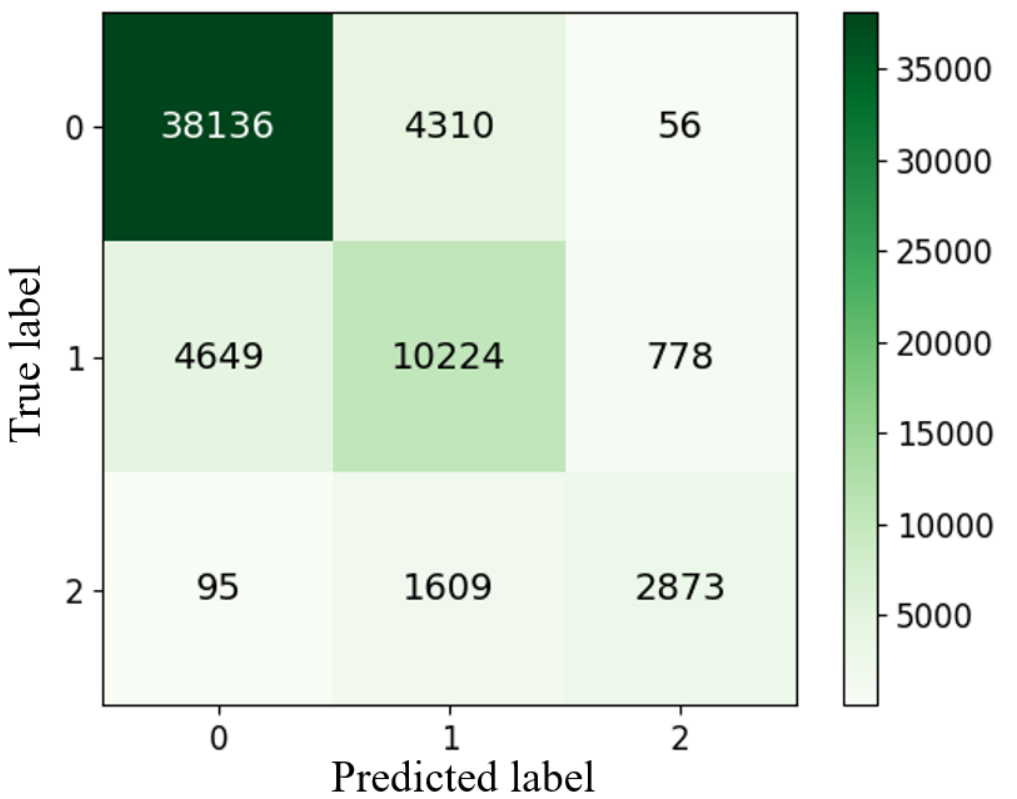}
	\caption{Confusion matrix of AI-Truck.}
	\label{fig11}
\end{figure}

\noindent Table \ref{Tab1} reports the overall performance of AI-Truck and other baselines on Chengdu's slag truck datasets. The mean and standard deviation of the 10 experiments are represented by values in and out of parentheses, respectively. The best results are highlighted in bold and the second best results are underlined. We can clearly see that our approach consistently outperforms all the baselines over most metrics, demonstrating the superiority of AI-Truck. Specifically, AI-Truck improves the macro F1 by $0.007$ compared to the state-of-the-art model (PDFormer). Figure \ref{fig11} displays the confusion matrix of AI-Truck, where Class 0, Class 1, and Class 2 represent the level of slag truck activity $c_{s}=0$, $c_{s}=1$, and $c_{s}=2$ respectively, highlighting AI-Truck has a super competence to predict the levels of slag truck activity.
 
\begin{table*}[htb]
	\centering
	\caption{Performance of Different Models.}
	\renewcommand{\baselinestretch}{1.2}
	\setlength{\tabcolsep}{2.5pt}
	\label{Tab1}
	\footnotesize
	\begin{tabular}{ccccccccccc}
		\hline
		\multirow{2}{*}{Model} & \multicolumn{3}{c}{Precision} & \multicolumn{3}{c}{Recall} & \multicolumn{3}{c}{F1 Score} & \multirow{2}{*}{\begin{tabular}[c]{@{}c@{}}Macro \\ F1\end{tabular}} 
		\\ \cline{2-10}
		& $c_s=2$ & $c_s=1$ & $c_s=0$ & $c_s=2$ & $c_s=1$ & $c_s=0$ & $c_s=2$ & $c_s=1$ & $c_s=0$ &                       \\ \cline{1-1} \cline{11-11} 
		\hline
		AI-Truck & \textbf{0.757}(\textbf{0.015}) & \textbf{0.615}(\textbf{0.006}) & $\underline{0.905}$(\textbf{0.005}) & $\underline{0.647}$(\textbf{0.017})& $\underline{0.706}$(\textbf{0.007})& $\underline{0.869}$(\textbf{0.006}) & \textbf{0.697}(\textbf{0.007}) & \textbf{0.657}(\textbf{0.005})& \textbf{0.886}(\textbf{0.004})& \textbf{0.747}(\textbf{0.002})               
		\\
		Bi-LSTM & 0.747(0.031)& $\underline{0.612}$(0.028) & 0.901(\textbf{0.005}) & 0.633(0.049) & 0.696(0.019) & 0.864($\underline{0.009}$) & 0.683(0.017) & 0.646($\underline{0.006}$) & 0.882($\underline{0.005}$) & 0.737($\underline{0.005}$)                                                \\
		TCN & $\underline{0.756}$($\underline{0.022}$) & 0.608(\textbf{0.006}) & 0.899(\textbf{0.005}) & 0.611($\underline{0.030}$) & 0.693($\underline{0.011}$) & \textbf{0.870}(0.010) & 0.675($\underline{0.013}$) & 0.648($\underline{0.006}$) & $\underline{0.884}$($\underline{0.005}$) & 0.736($\underline{0.005}$)                            
		\\
		STGCN & 0.677(0.027) & 0.575(0.030) & 0.885($\underline{0.011}$) & 0.572(0.035) & 0.657(0.032) & 0.844(0.025) & 0.619(0.014) & 0.616(0.026) & 0.864(0.009) & 0.696($\underline{0.005}$)                                   
		\\
		PDFormer & 0.737(0.024) & 0.599($\underline{0.009}$) & \textbf{0.907}(\textbf{0.005}) & \textbf{0.650}(0.047) & \textbf{0.709}(0.016) & 0.856(0.012) & $\underline{0.689}$(0.018) & $\underline{0.649}$(\textbf{0.005}) & 0.881(0.006) & $\underline{0.740}$($\underline{0.005}$)                                      
		\\ 
		\hline
	\end{tabular}
\end{table*}

\subsection{Ablation Study}
\subsubsection{Effects of Bagging} AI-Truck improves its stability by integrating multiple independent base models with poor stability through a soft vote integrator, which is called bagging. It can be seen from Table \ref{Tab1} that the standard deviation of macro F1 is 0.003 lower than the base models with the most stable prediction results, fully demonstrating the effectiveness of bagging. 
\subsubsection{Effects of Downsampling and Weighted Loss}
In this paper, we tackle the challenge of imbalanced spatial distribution by applying the combination of downsampling and weighted loss. This combination could improve the model’s performance for the minority classes ($c_{s}=1$ and $c_{s}=2$) while preserving satisfactory performance for the majority class ($c_{s}=0$). Downsampling reduces the number of majority class samples, thereby enabling the model to allocate greater attention to the minority classes. While weighted loss approach increases the impact of minority classes on the model's loss function during training by assigning them higher loss weights. Table \ref{Tab2} reports the comparative overall performance of AI-Truck and PDFormer (the best base model) when trained with different methods aimed to address the issue of imbalanced spatial distribution, demonstrating the combination of downsampling and weighted loss significantly ameliorates the effects of data imbalance. To be specific, our method significantly improves the F1 scores for $c_{s}=2$ and $c_{s}=1$, with increases of $0.102$ and $0.036$, respectively, for AI-Truck, and $0.101$ and $0.073$ for PDFormer, compared to the models without using any methods. Moreover, downsampling is more effective than weighted loss in improving the model's prediction for the minority classes according to Table \ref{Tab2}.
\begin{table*}[htb]
	\centering
	\caption{Performance of Different methods to address the issue of imbalanced spatial distribution}
	\renewcommand{\baselinestretch}{1.2}
	\setlength{\tabcolsep}{1.2pt}
	\label{Tab2}
	\footnotesize
	\begin{tabular}{cccccccccccc}
		\hline
		\multirow{2}{*}{Model} & \multirow{2}{*}{Method} & \multicolumn{3}{c}{Precision} & \multicolumn{3}{c}{Recall} & \multicolumn{3}{c}{F1 Score} & \multirow{2}{*}{\begin{tabular}[c]{@{}c@{}}Macro \\ F1\end{tabular}} \\ \cline{3-11}
		&  & $c_s=2$ & $c_s=1$ & $c_s=0$ & $c_s=2$ & $c_s=1$ & $c_s=0$ & $c_s=2$ & $c_s=1$ & $c_s=0$ &  \\
		\hline
		\multirow{4}{*}{AI-Truck} & \begin{tabular}[c]{@{}c@{}}down-sampling and \\ weighted loss\end{tabular} & 0.757 & 0.615 & 0.905 & \textbf{0.647} & \textbf{0.706} & 0.869 & $\underline{0.697}$ & \textbf{0.657} & 0.886 & \textbf{0.747}\\
		& down-sampling & 0.760 & \textbf{0.656} & 0.875 & $\underline{0.645}$ & 0.597 & 0.919 & \textbf{0.698} & $\underline{0.625}$ & 0.897 & 0.740 \\
		& weighted loss & \textbf{0.794} & 0.590 & \textbf{0.967} & 0.557 & $\underline{0.617}$ & $\underline{0.968}$ & 0.655 & 0.603 & $\underline{0.967}$ & $\underline{0.742}$ \\& 
		none & $\underline{0.777}$ & $\underline{0.653}$ & $\underline{0.956}$ & 0.576 & 0.482 & \textbf{0.983} & 0.661 & 0.555 & \textbf{0.969} & 0.728 
		\\ \cline{2-12}
		\multirow{4}{*}{PDformer} & 
		\begin{tabular}[c]{@{}c@{}}down-sampling and \\ weighted loss\end{tabular} & 0.737 & 0.599 & 0.907 & \textbf{0.650} & \textbf{0.709} & 0.856 & \textbf{0.689} &\textbf{0.649} & 0.881 & \textbf{0.740}  \\
	    & down-sampling & $\underline{0.768}$ & \textbf{0.646} & 0.881 & $\underline{0.597}$ & $\underline{0.615}$ & 0.916 & $\underline{0.672}$ & $\underline{0.630}$ & 0.898 & $\underline{0.733}$ \\
		& weighted loss & 0.669 & 0.541 & \textbf{0.962} & 0.576 & 0.571 & $\underline{0.960}$ & 0.619 & 0.555 & $\underline{0.961}$ & 0.712 \\
		& none & \textbf{0.781} & $\underline{0.616}$ & $\underline{0.954}$ & 0.509 & 0.493 & \textbf{0.979} & 0.616 & 0.548 & \textbf{0.966} & 0.710 
		\\ \hline
	\end{tabular}
\end{table*}

\subsubsection{Effects of the Geographic Feature base on both OD and neighboring relationship} 
To validate the effectiveness of the geographic feature proposed in this paper, we perform a comparison with geographic feature from other researches that only based on the neighboring relationship. Their performances are reported in Table \ref{Tab3}. Notably, the geographic feature in this paper significantly outperforms those from previous researches. Specifically, it improves macro F1 by 0.007 and 0.010 on AI-Truck and PDFormer respectively. This improvement can be attributed to the geographic feature base on both OD and neighboring relationship is closer alignment with real-world conditions.

\begin{table*}[htb]
	\centering
	\caption{Performance of Different Geographic Features.}
	\renewcommand{\baselinestretch}{1.2}
	\setlength{\tabcolsep}{1.2pt}
	\label{Tab3}
	\footnotesize
	\begin{tabular}{cccccccccccc}
		\hline
		\multirow{2}{*}{Model} & \multirow{2}{*}{\begin{tabular}[c]{@{}c@{}}Geographic \\ Features\end{tabular}} & \multicolumn{3}{c}{Precision} & \multicolumn{3}{c}{Recall} & \multicolumn{3}{c}{F1 Score} & \multirow{2}{*}{\begin{tabular}[c]{@{}c@{}}Macro \\ F1\end{tabular}} \\ \cline{3-11}
		&  & $c_s=2$ & $c_s=1$ & $c_s=0$ & $c_s=2$ & $c_s=1$ & $c_s=0$ & $c_s=2$ & $c_s=1$ & $c_s=0$ &  \\
		\hline
		\multirow{2}{*}{AI-Truck} & OD and neighboring relationship & \textbf{0.757} & \textbf{0.615} & 0.905 & \textbf{0.647} & 0.706 & \textbf{0.869} & \textbf{0.697} & 0.657 & \textbf{0.886} & \textbf{0.747}\\& Neighboring relationship & 0.751 & 0.600 & \textbf{0.909} & 0.627 & \textbf{0.729} & 0.851 & 0.684 & \textbf{0.658} & 0.879 & 0.740
		\\ \cline{2-12}
		\multirow{2}{*}{PDFormer} & OD and neighboring relationship & 0.737 & \textbf{0.599} & 0.907 & \textbf{0.650} & 0.709 & \textbf{0.856} & \textbf{0.689} & \textbf{0.649} & \textbf{0.881} & \textbf{0.740}\\& Neighboring relationship & \textbf{0.754} & 0.583 & \textbf{0.911} & 0.595 & \textbf{0.721} & 0.854 & 0.665 & 0.645 & \textbf{0.881} & 0.730
		\\ \hline
	\end{tabular}
\end{table*}

\begin{figure}[htb]
	\centering
	\includegraphics[width=.45\textwidth]{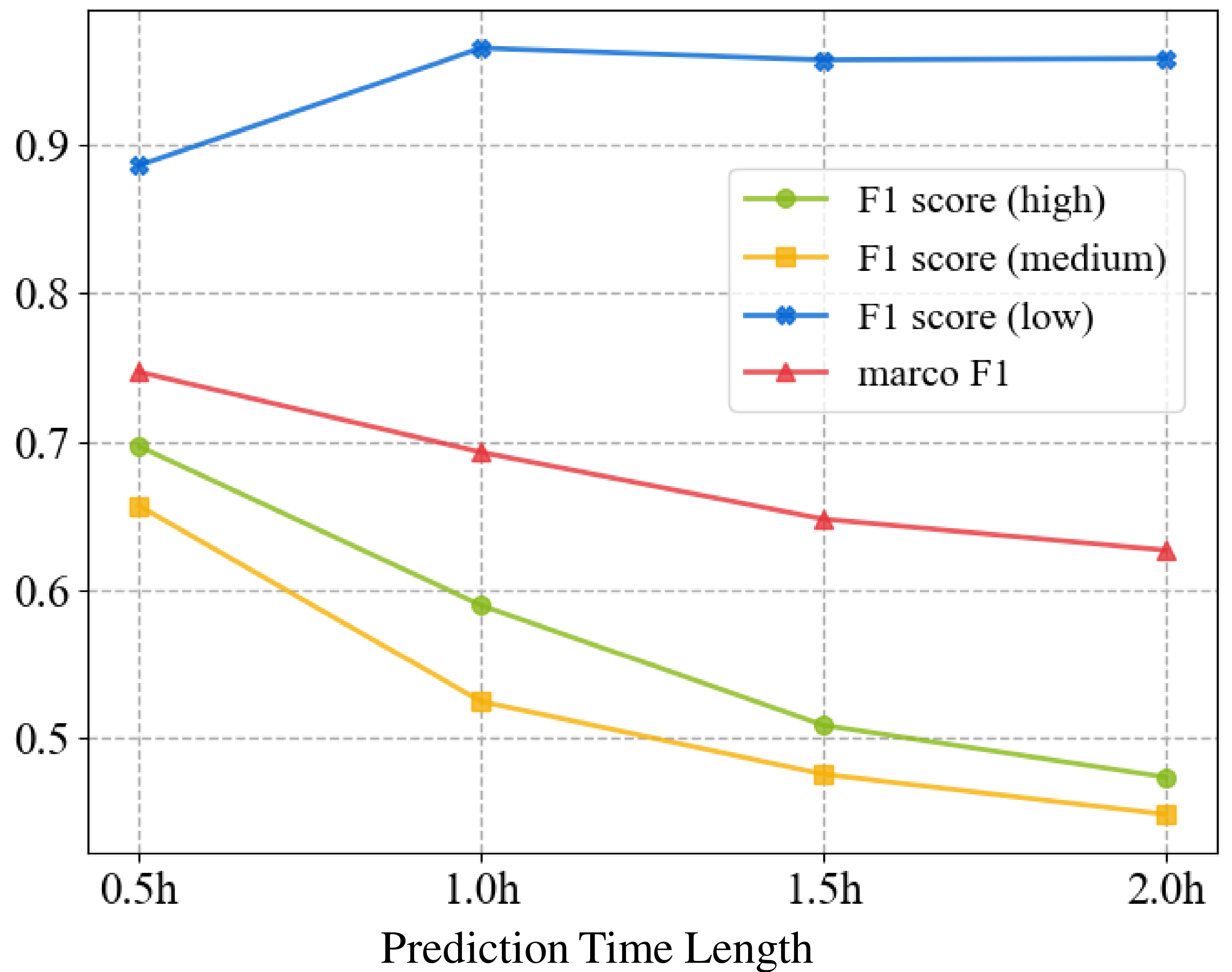}
	\caption{F1 Scores and Macro F1 of AI-Truck on Different Prediction Time Lengths.}
	\label{F1_time}
\end{figure} 

\subsection{Hyperparameter Study}
\noindent To explore the impact of prediction time length on AI-Truck's performance, we set output windows to values ranging from 1 to 4, corresponding to prediction time lengths of 0.5 to 2.0-h. Figure \ref{F1_time} illustrates the overall performance of AI-Truck across different prediction time lengths. With the increasing prediction time length, the F1 scores for high and medium level of slag truck activity, along with the macro F1 decline, while the F1 score for low level of slag truck activity increases, which indicates that AI-truck's performance worsens with longer prediction time lengths, and the impact of imbalanced spatial distribution also becomes more severe. Fortunately, it is noteworthy that the decrease in AI-Truck's macro F1 is gradual, maintaining a score above 0.6 even at a 2.0-h prediction time length.

\begin{figure*}[htb]
	\centering
	\includegraphics[width=.9\textwidth]{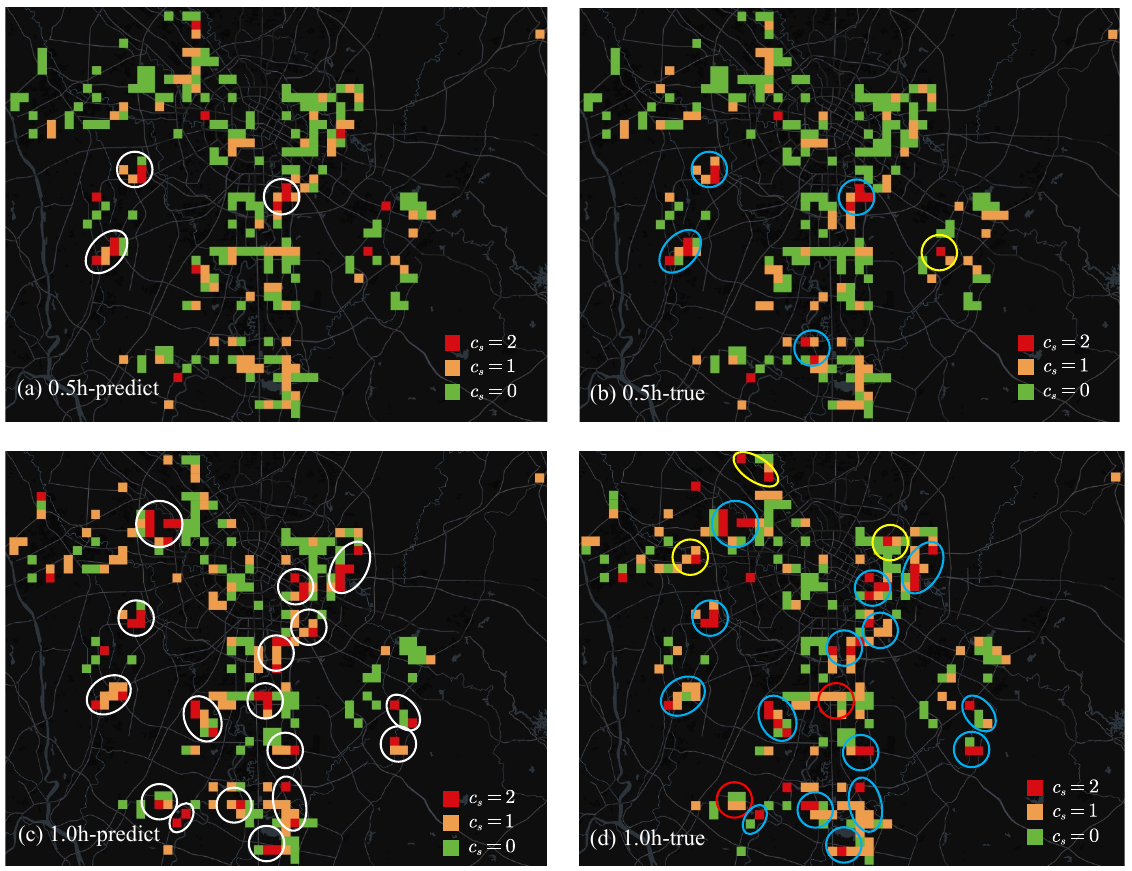}
	\caption{Predicted and true levels of slag truck activity for 0.5-h and 1.0-h prediction in the case study. White circles in (a) and (c) mark high-activity areas identified by AI-Truck; blue (or red) circles in (b) and (d) means the predictions are correct (or incorrect) in an approximate sense; yellow circles indicate areas missed by AI-Truck.}
	\label{figaccuracy}
\end{figure*}

\subsection{Case study}

In this section, we conduct a case study of AI-Truck by linking its prediction capability to real-world operations. Figure \ref{figaccuracy} visualizes the predicted and true classes of individual grids for 0.5-h and 1.0-h prediction. In actual law enforcement operations, personnel tend to focus on areas where several high-activity ($c_s=2$) grids are clustered, as highlighted by white circles in Figure \ref{figaccuracy} (a) and (c). In practice, these predictions are considered accurate if one or more high-activity ($c_s=2$) grids are present in the vicinity, as indicated by the blue circles. We adopt such a loose definition of accuracy because even if the prediction is not accurate for each individual grids, the personnel can quickly move to near-by grids with high truck activities. Following this criterion, the 3 locations predicted in subfigure (a) are all accurate, and 15 out of the 17 locations in subfigure (c) are accurate. Moreover, the yellow circles in subfigures (b) and (d) mark those high-activity locations not identified by AI-Truck. But these numbers are small. Overall, the proposed method shows superior performance in predicting areas with high truck activity, provided minor spatial inaccuracies can be tolerated. 

\begin{figure}[htb]
	\centering
	\includegraphics[width=0.45\textwidth]{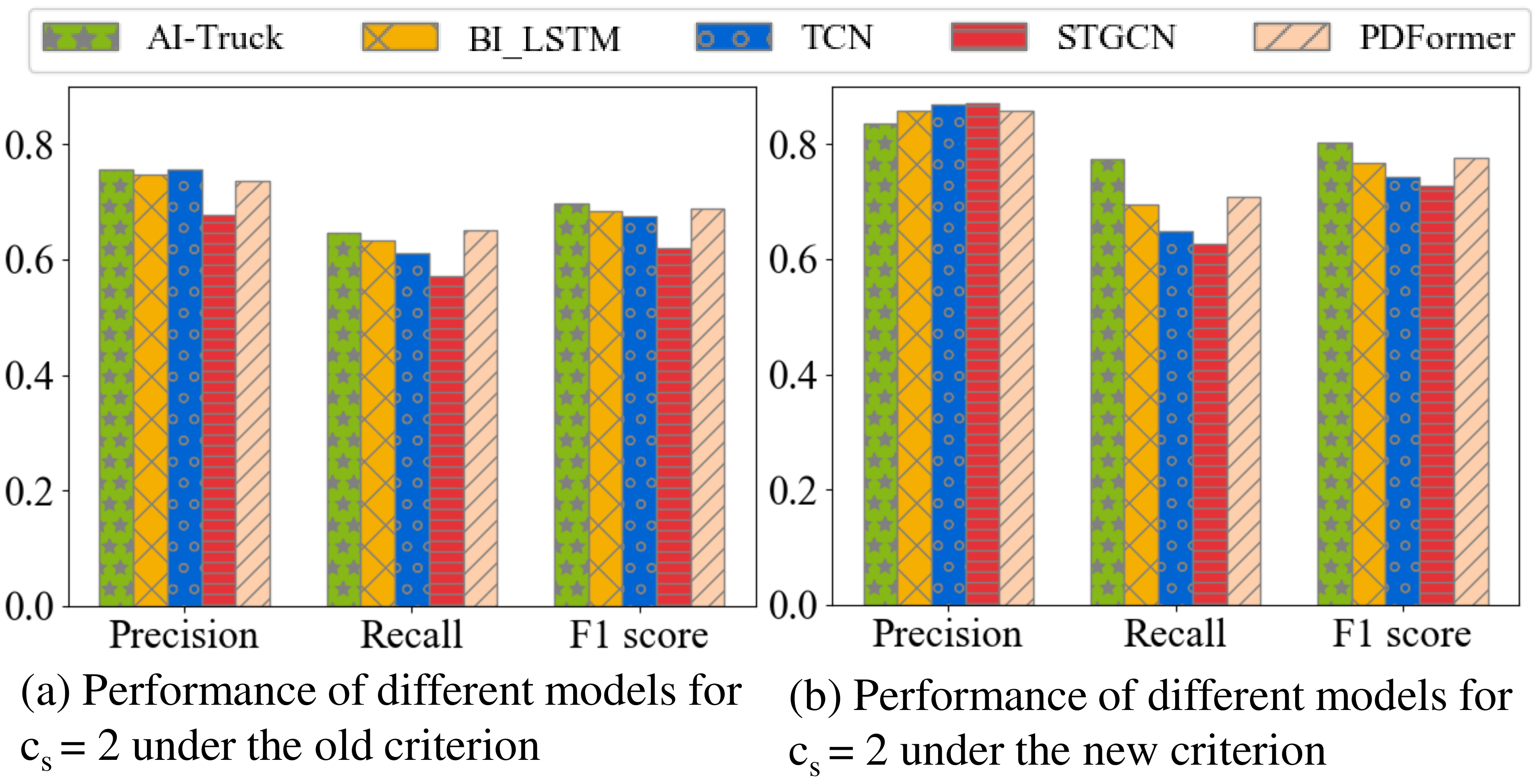}
	\caption{Performance of different models for predicting high-activity locations under different criteria.}
	\label{high_level}
\end{figure}

To further validate the above observation, we test the model performances (precision, recall, F1 score) under the following revised definition: A grid $s$ is deemed to have high truck activities, if and only if $c_s=2$ or any of its one-hop neighboring grids $s'$ has $c_{s'}=2$. Figure \ref{high_level} reports the precision, recall and F1 score for predicting high-activity grids under the old and new criteria. Understandably, all the five methods have improved performance, and AI-Truck leads in recall and F1 score. 
 
\begin{figure}[htb]
	\centering
	\includegraphics[width=0.45\textwidth]{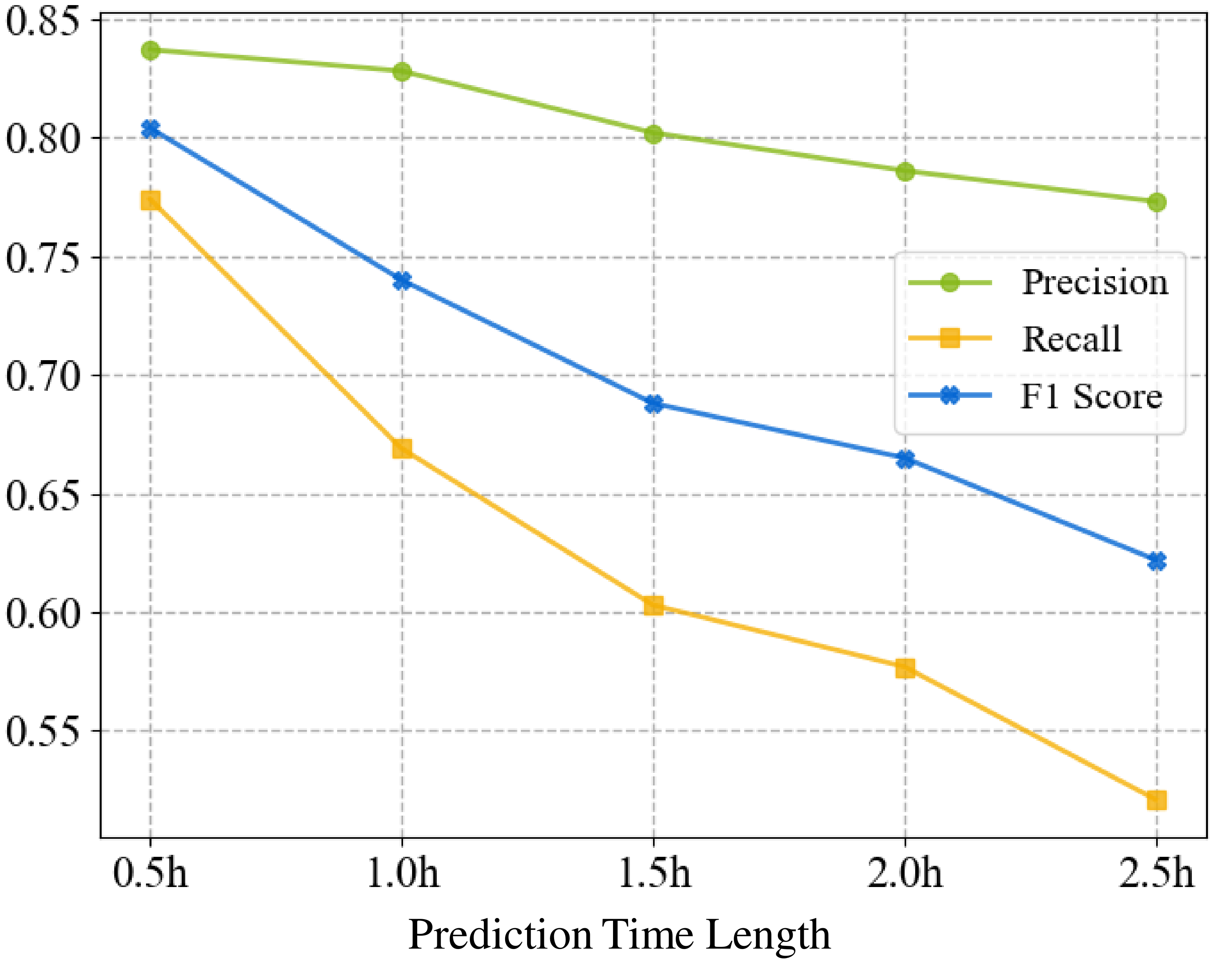}
	\caption{Performance of predicting high-activity locations for different prediction time lengths.}
	\label{fig12}
\end{figure}

As shown in Figure \ref{fig12}, the precision, recall and F1 score of AI-Truck for predicting high-activity locations decline as the prediction time length increases, but the precision is less impacted. In particular, the precision is above 0.8 for 1.5h-prediction, meaning that the personnel have abundant time to reach relevant areas and spot high-level truck activities, 4 out of 5 times, using AI-Truck. 

\section{Conclusion}
\noindent In this paper, we propose AI-Truck, a novel deep ensemble framework for predicting the levels of slag truck activity. The core strength of AI-Truck is its bagging strategy, which improves both the stability and accuracy of predictions. In order to address the issue of imbalanced spatial distribution, we employ the combination of downsampling and weighted loss in the data processing and training phases respectively. In terms of geographic feature, we use slag truck trajectory points to identify OD relationship between neighboring grids, which is a departure from previous studies. Extensive experiments on Chengdu's slag truck datasets demonstrate that our AI-Truck consistently outperforms previous state-of-the-art methods, achieving a macro F1 of $0.747$. Moreover, we visualize AI-Truck's prediction and propose a loose definition of accuracy, which is better suited to the real scenario. As part of our future work, we plan to continue enhance AI-Truck by integrating additional base models and extracting a broader range of spatio-temporal features to further improve the accuracy and stability of our predictions.

	\begin{IEEEbiography}[{\includegraphics[width=1in,height=1.25in,clip,keepaspectratio]{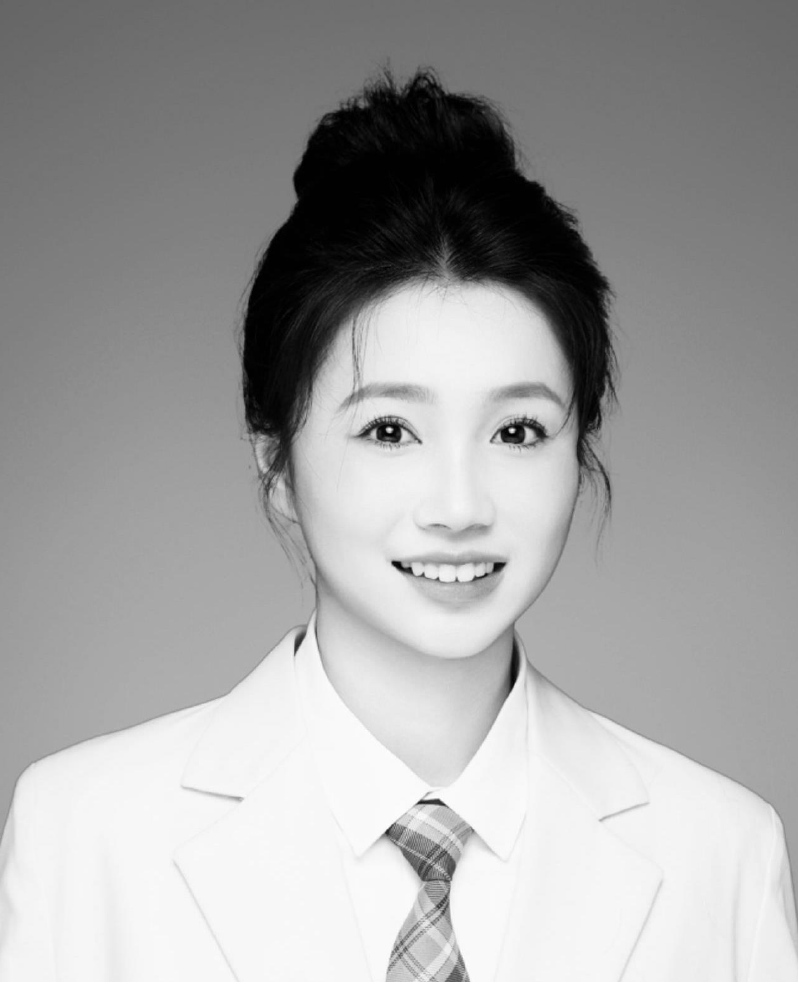}}]{Meng Xu} obtained the B.S. degree in transportation from Southwest Jiaotong University, Chengdu, China, in 2023. She is currently pursuing the M.S. degree with the School of Transportation and Logistics at Southwest Jiaotong University, Chengdu, China. Her research interests lie in urban computing and spatio-temporal data mining.
	\end{IEEEbiography}
	
	\begin{IEEEbiography}[{\includegraphics[width=1in,height=1.25in,clip,keepaspectratio]{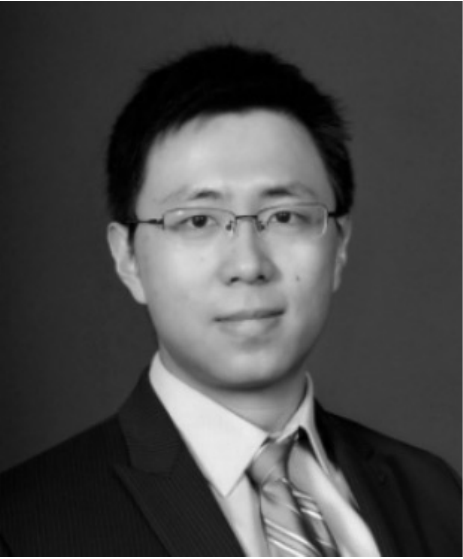}}]{Ke Han}
	received the B.S. degree in mathematics from the University of Science and Technology of China in 2008, and the Ph.D. degree in mathematics from Pennsylvania State University in 2013. He was appointed as a Lecturer in 2013, and then a Senior Lecturer in 2018, with the Centre for Transport Studies, Department of Civil and Environmental Engineering, Imperial College London. He joined Southwest Jiaotong University, China, as a Full Professor in 2020. He has published two book and over 140 journal and conference papers in the areas of transportation modeling and optimization, and sustainable urban management.

	\end{IEEEbiography}
	
		\vfill


\begin{thebibliography}{1}
\bibliographystyle{IEEEtran}

\bibitem{ref1}
M. Kampa and E. Castanas, “Human health effects of air pollution,”{\it{Environmental polllution}}, vol. 151, no. 2, pp. 362–367, 2008.

\bibitem{ref2}
B. Brunekreef and S. T. Holgate, “Air pollution and health,” {\it{The lancet}}, vol. 360, no. 9341, pp. 1233–1242, 2002.

\bibitem{ref3}
J. A. Bernstein, N. Alexis, C. Barnes, I. L. Bernstein, A. Nel, D. Peden, D. Diaz-Sanchez, S. M. Tarlo, and P. B. Williams, “Health effects of air pollution,” {\it{Journal of allergy and clinical immunology}}, vol. 114, no. 5, pp. 1116–1123, 2004.

\bibitem{refnew1}
PASTORELLO, C. \& MELIOS, G. 2016. Explaining road transport emissions: a non-technical guide.


\bibitem{ref4}
X. Deng, “Economic costs of motor vehicle emissions in china: a case study,” {\it{Transportation Research Part D: Transport and Environment}}, vol. 11, no. 3, pp. 216–226, 2006.

\bibitem{refMG2015}
K. Milton and J. Grix, “Public health policy and walking in england—analysis of the 2008 ‘policy window’,” BMC public health, vol. 15, no. 1, pp. 1–9, 2015.

\bibitem{refSMMSP2022}
F. da Silva Souza, J. C. Mendes, L. J. B. Morais, J. S. Silva, and R. A. F. Peixoto, “Mapping and recycling proposal for the construction and demolition waste generated in the brazilian amazon,” Resources, Conservation and Recycling, vol. 176, p. 105896, 2022.

\bibitem{refMBS2021}
L. M. Maués, N. Beltrão, and I. Silva, “Ghg emissions assessment of civil construction waste disposal and transportation process in the eastern amazon,” Sustainability, vol. 13, no. 10, p. 5666, 2021.

\bibitem{ref5}
F. Karagulian, C. A. Belis, C. F. C. Dora, A. M. Prüss-Ustün, S. Bonjour, H. AdairRohani, and M. Amann, “Contributions to cities’ ambient particulate matter (pm): A systematic review of local source contributions at global level,” {\it{Atmospheric environment}}, vol. 120, pp. 475–483, 2015.

\bibitem{ref6}
P. Jiang, X. Zhong, and L. Li, “On-road vehicle emission inventory and its spatio-temporal variations in north china plain,” {\it{Environmental Pollution}}, vol. 267, p. 115639, 2020.


\bibitem{ref65} Ministry of Ecology and Environment of China, ``China mobile source environmental management annual report", 2022.

\bibitem{ref8}
Yukun Ma, Manli Gong, Hongtao Zhao, and Xuyong Li. Contribution of road dust from low impact development (lid) construction sites to atmospheric pollution from heavy metals. Science of The Total Environment, 698:134243, 2020.

\bibitem{ref9}
H. Yan, G. Ding, K. Feng, L. Zhang, H. Li, Y. Wang, and T. Wu, “Systematic evalua-tion framework and empirical study of the impacts of building construction dust on the surrounding environment,” Journal of Cleaner Production, vol. 275, p. 122767, 2020.

\bibitem{ref10}
S. Yang, J. Liu, X. Bi, Y. Ning, S. Qiao, Q. Yu, and J. Zhang, “Risks related to heavy metal pollution in urban construction dust fall of fast-developing chinese cities,” Ecotoxicology and Environmental Safety, vol. 197, p. 110628, 2020.

\bibitem{ref7}
Haibin Wang, Lihui Han, Tingting Li, Song Qu, Yuncheng Zhao, Shoubin Fan, Tong Chen, Haoran Cui, and Junfang Liu. Temporal-spatial distributions of road silt loadings and fugitive road dust emissions in beijing from 2019 to 2020. Journal of Environmental Sciences, 132:56–70, 2023.

\bibitem{PTG2016} Perez-Torres, R., Torres-Huitzil, C., \& Galeana-Zapien, H. (2016). Full On-Device Stay Points Detection in Smartphones for Location-Based Mobile Applications. Sensors (Basel), 16(10)

\bibitem{refWYYBL2022}
X. Wei, M. Ye, L. Yuan, W. Bi, and W. Lu, “Analyzing the freight characteristics and carbon emission of construction waste hauling trucks: big data analytics of hong kong,” International Journal of Environmental Research and Public Health, vol. 19, no. 4, p. 2318, 2022.

\bibitem{refL2019}
W. Lu, “Big data analytics to identify illegal construction waste dumping: A hong kong study,” Resources, conservation and recycling, vol. 141, pp. 264–272, 2019.

\bibitem{refLYL2022}
W. Lu, L. Yuan, and W. M. Lee, “Understanding loading patterns of construction waste hauling trucks: triangulation between big quantitative and informative qualitative data,” Environmental Science and Pollution Research, vol. 29, no. 33, pp. 50 867–50 880, 2022.

\bibitem{refLK2024}L. Yu and K. Han, “Using construction waste hauling trucks’ GPS data to classify earthwork-related locations: A Chengdu case study,” arXiv preprint arXiv:2402.14698, 2024.

\bibitem{ref12}
D. A. Tedjopurnomo, Z. Bao, B. Zheng, F. M. Choudhury, and A. K. Qin, “A survey
on modern deep neural network for traffic prediction: Trends, methods and challenges,” {\it{IEEE Transactions on Knowledge and Data Engineering}}, vol. 34, no. 4, pp. 1544–1561, 2020.

\bibitem{ref13}
W. Huang, G. Song, H. Hong, and K. Xie, “Deep architecture for traffic flow prediction: Deep belief networks with multitask learning,” {\it{IEEE Transactions on Intelligent Transportation Systems}}, vol. 15, no. 5, pp. 2191–2201, 2014.

\bibitem{ref14}
H. Yu, Z. Wu, S. Wang, Y. Wang, and X. Ma, “Spatiotemporal recurrent convolutional networks for traffic prediction in transportation networks,” {\it{Sensors}}, vol. 17, no. 7, p. 1501, 2017.

 \bibitem{ref16}
H. Yao, F. Wu, J. Ke, X. Tang, Y. Jia, S. Lu, P. Gong, J. Ye, and Z. Li, “Deep multiview spatial-temporal network for taxi demand prediction,” in {\it{Proceedings of the AAAI conference on artificial intelligence}}, vol. 32, no. 1, 2018.

 \bibitem{ref19}
B. Yu, H. Yin, and Z. Zhu, “Spatio-temporal graph convolutional networks: A deep learning framework for traffic forecasting,” {\it{arXiv preprint arXiv:1709.04875}}, 2017.

\bibitem{refQBY2018}
Q. Liu, B. Wang, and Y. Zhu, “Short-term traffic speed forecasting based on attention convolutional neural network for arterials,” Computer-Aided Civil and Infrastructure Engineering, vol. 33, no. 11, pp. 999-1016, 2018.

\bibitem{refYJY2016}
Y. Jia, J. Wu, and Y. Du, “Traffic speed prediction using deep learning method,” in 2016 IEEE 19th International Conference on Intelligent Transportation Systems (ITSC), pp. 1217-1222, 2016.

 \bibitem{ref22}
Z. Fang, Q. Long, G. Song, and K. Xie, “Spatial-temporal graph ode networks for traffic flow forecasting,” in {\it{Proceedings of the 27th ACM SIGKDD conference on knowledge discovery \& data mining}}, 2021, pp. 364–373.

\bibitem{refYLSDY2017}
R. Yu, Y. Li, C. Shahabi, U. Demiryurek, and Y. Liu, “Deep learning: A generic approach for extreme condition traffic forecasting,” in Proceedings of the 2017 SIAM international Conference on Data Mining. SIAM, 2017, pp. 777-785.

\bibitem{refXJ2020}
H. Xu and C. Jiang, “Deep belief network-based support vector regression method for traffic flow forecasting,” Neural Computing and Applications, vol. 32, pp. 2027-2036, 2020.

\bibitem{ref28}
P. Bühlmann and B. Yu, “Analyzing bagging,” {\it{The annals of Statistics}}, vol. 30, no. 4, pp. 927–961, 2002.

\bibitem{ref29}
Z.-H. Zhou, {\it{Machine learning}}. Springer Nature, 2021.

 \bibitem{ref23}
J. Choi, H. Choi, J. Hwang, and N. Park, “Graph neural controlled differential equations for traffic forecasting,” in {\it{Proceedings of the AAAI Conference on Artificial Intelligence}}, vol. 36, no. 6, 2022, pp. 6367–6374.

 \bibitem{ref24}
Y. Li, R. Yu, C. Shahabi, and Y. Liu, “Diffusion convolutional recurrent neural network: Data-driven traffic forecasting,” {\it{arXiv preprint arXiv:1707.01926}}, 2017.

\bibitem{ref25}
S. Guo, Y. Lin, N. Feng, C. Song, and H. Wan, “Attention based spatial-temporal graph convolutional networks for traffic flow forecasting,” in{\it{Proceedings of the AAAI conference on artificial intelligence}} , vol. 33, no. 01, 2019, pp. 922–929.

\bibitem{ref38}
M. Xu, W. Dai, C. Liu, X. Gao, W. Lin, G.-J. Qi, and H. Xiong, “Spatial-temporal transformer networks for traffic flow forecasting,” {\it{arXiv preprint arXiv:2001.02908}}, 2020.

\bibitem{ref40}
S. Yang, S. Shi, X. Hu, and M. Wang, “Spatiotemporal context awareness for urban traffic modeling and prediction: sparse representation based variable selection,” {\it{PloS one}}, vol. 10, no. 10, p. e0141223, 2015.

\bibitem{ref39}
A. Ali, Y. Zhu, and M. Zakarya, “Exploiting dynamic spatio-temporal graph convolutional neural networks for citywide traffic flows prediction,” {\it{Neural networks}}, vol. 145, pp. 233–247, 2022.

\bibitem{ref15}
J. Zhang, Y. Zheng, and D. Qi, “Deep spatio-temporal residual networks for citywide crowd flows prediction,” in {\it{Proceedings of the AAAI conference on artificial intelligence}}, vol. 31, no. 1, 2017.

 \bibitem{ref26}
J. Jiang, C. Han, W. X. Zhao, and J. Wang, “Pdformer: Propagation delay-aware dynamic long-range transformer for traffic flow prediction,” {\it{arXiv preprint arXiv:2301.07945}}, 2023.

\bibitem{ref41}
J. Zhang, Y. Zheng, D. Qi, R. Li, and X. Yi, “Dnn-based prediction model for spatio-temporal data,” in {\it{Proceedings of the 24th ACM SIGSPATIAL international conference on advances in geographic information systems}}, 2016, pp. 1–4.

\bibitem{ref42}
S. Salvador and P. Chan, “Toward accurate dynamic time warping in linear time and space,” {\it{Intelligent Data Analysis}}, vol. 11, no. 5, pp. 561–580, 2007.

\bibitem{ref45}
Y. Wu, H. Tan, L. Qin, B. Ran, and Z. Jiang, “A hybrid deep learning based traffic flow prediction method and its understanding,” {\it{Transportation Research Part C: Emerging Technologies}}, vol. 90, pp. 166–180, 2018.

\bibitem{ref46}
H. Tan, G. Feng, J. Feng, W. Wang, Y.-J. Zhang, and F. Li, “A tensor-based method for missing traffic data completion,” {\it{Transportation Research Part C: Emerging Technologies}}, vol. 28, pp. 15–27, 2013.

\bibitem{ref47}
T. Thianniwet, S. Phosaard, and W. Pattara-Atikom, “Classification of road traffic congestion levels from gps data using a decision tree algorithm and sliding windows,” in {\it{Proceedings of the world congress on engineering}}, vol. 1, 2009, pp. 1–3.

\bibitem{ref30}
L. I. Kuncheva and C. J. Whitaker, “Measures of diversity in classifier ensembles and their relationship with the ensemble accuracy,” {\it{Machine learning}}, vol. 51, pp. 181–207, 2003.

\bibitem{refMK1997}
M. Schuster and K. K. Paliwal, “Bidirectional recurrent neural networks,” IEEE Transactions on Signal Processing, vol. 45, no. 11, pp. 2673-2681, 1997.

\bibitem{refCMRAG2017}
C. Lea, M. D. Flynn, R. Vidal, A. Reiter, and G. D. Hager, “Temporal convolutional networks for action segmentation and detection,” in Proceedings of the IEEE Conference on Computer Vision and Pattern Recognition, pp. 156-165, 2017.




































\bibitem{ref11}
X. Yin, G. Wu, J. Wei, Y. Shen, H. Qi, and B. Yin, “Deep learning on traffic prediction: Methods, analysis, and future directions,” {\it{IEEE Transactions on Intelligent Transportation Systems}}, vol. 23, no. 6, pp. 4927–4943, 2021.




 


 \bibitem{ref17}
 A. Borovykh, S. Bohte, and C. W. Oosterlee, “Conditional time series forecasting with convolutional neural networks,” {\it{arXiv preprint arXiv:1703.04691}}, 2017.

 \bibitem{ref18}
 S. Bai, J. Z. Kolter, and V. Koltun, “An empirical evaluation of generic convolutional and recurrent networks for sequence modeling,” {\it{arXiv preprint arXiv:1803.01271}}, 2018.



 \bibitem{ref20}
 Z. Wu, S. Pan, G. Long, J. Jiang, X. Chang, and C. Zhang, “Connecting the dots: Multivariate time series forecasting with graph neural networks,” in {\it{Proceedings of the 26th ACM SIGKDD international conference on knowledge discovery \& data mining}}, 2020, pp. 753–763.

 \bibitem{ref21}
 Z. Wu, S. Pan, G. Long, J. Jiang, and C. Zhang, “Graph wavenet for deep spatial-temporal graph modeling,” {\it{arXiv preprint arXiv:1906.00121}}, 2019.





 \bibitem{ref27}
 W. Chen, L. Chen, Y. Xie, W. Cao, Y. Gao, and X. Feng, “Multi-range attentive bicomponent graph convolutional network for traffic forecasting,” in {\it{Proceedings of the AAAI conference on artificial intelligence}} 
, vol. 34, no. 04, 2020, pp. 3529–3536.





\bibitem{ref31}
H. B. Mitchell and P. A. Schaefer, “A “soft” k-nearest neighbor voting scheme,” {\it{International journal of intelligent systems}}, vol. 16, no. 4, pp. 459–468, 2001.

\bibitem{ref32}
Z. Xu, M. Bashir, W. Zhang, Y. Yang, X. Wang, and C. Li, “An intelligent fault diagnosis for machine maintenance using weighted soft-voting rule based multi-attention module with multi-scale information fusion,” {\it{Information Fusion}}, vol. 86, pp. 17–29, 2022.

\bibitem{ref33}
Q. Zhou and H. Wu, “Nlp at iest 2018: Bilstm-attention and lstm-attention via soft voting in emotion classification,” in {\it{Proceedings of the 9th workshop on computational approaches to subjectivity}}, sentiment and social media analysis, 2018, pp. 189–194.

\bibitem{ref34}
S. Kumari, D. Kumar, and M. Mittal, “An ensemble approach for classification and prediction of diabetes mellitus using soft voting classifier,” {\it{International Journal of Cognitive Computing in Engineering}}, vol. 2, pp. 40–46, 2021.

\bibitem{ref35}
E. I. Vlahogianni, M. G. Karlaftis, and J. C. Golias, “Short-term traffic forecasting: Where we are and where we’re going,” {\it{Transportation Research Part C: Emerging Technologies}}, vol. 43, pp. 3-19, 2014.




\bibitem{refBreiman1996} L. Breiman, “Bagging predictors,” Machine learning, vol. 24, pp. 123-140, 1996.





\bibitem{ref36}
D. R. Gentner, D. R. Worton, G. Isaacman, L. C. Davis, T. R. Dallmann, E. C. Wood, S. C. Herndon, A. H. Goldstein, and R. A. Harley, “Chemical composition of gas-phase organic carbon emissions from motor vehicles and implications for ozone production,”{\it{Environmental Science \& Technology}}, vol. 47, no. 20, pp. 11 837–11 848, 2013

\bibitem{ref37}
P. Xie, T. Li, J. Liu, S. Du, X. Yang, and J. Zhang, “Urban flow prediction from spatiotemporal data using machine learning: A survey,” {\it{Information Fusion}}, vol. 59, pp. 1–12, 2020.









\bibitem{ref43}
T. Rakthanmanon, B. Campana, A. Mueen, G. Batista, B. Westover, Q. Zhu, J. Zakaria, and E. Keogh, “Searching and mining trillions of time series subsequences under dynamic time warping,” in {\it{Proceedings of the 18th ACM SIGKDD international conference on Knowledge discovery and data mining}}, 2012, pp. 262–270.

\bibitem{ref44}
O. Maimon and L. Rokach, {\it{Data mining and knowledge discovery handbook. Springer}}, 2005, vol. 2, no. 2005.







\bibitem{ref48}
R. L. Abduljabbar, H. Dia, and P.-W. Tsai, “Unidirectional and bidirectional lstm models for short-term traffic prediction,” {\it{Journal of Advanced Transportation}}, vol. 2021, pp. 1–16, 2021.

\bibitem{ref49}
 R. Zhang, F. Sun, Z. Song, X. Wang, Y. Du, and S. Dong, “Short-term traffic flow forecasting model based on ga-tcn,” {\it{Journal of Advanced Transportation}}, vol. 2021, pp. 1–13,2021.
 
\bibitem{ref50}
Z. Pan, Y. Liang, W. Wang, Y. Yu, Y. Zheng, and J. Zhang, “Urban traffic prediction from
spatio-temporal data using deep meta learning,” in Proceedings of the 25th ACM SIGKDD
international conference on knowledge discovery \& data mining, 2019, pp. 1720–1730.

\bibitem{ref51}
S. M. Abd Elrahman and A. Abraham, “A review of class imbalance problem,” Journal
of Network and Innovative Computing, vol. 1, no. 2013, pp. 332–340, 2013.

\bibitem{ref52}
S.-J. Yen and Y.-S. Lee, “Cluster-based under-sampling approaches for imbalanced data
distributions,” Expert Systems with Applications, vol. 36, no. 3, pp. 5718–5727, 2009.

\bibitem{ref53}
E. Chamseddine, N. Mansouri, M. Soui, and M. Abed, “Handling class imbalance in
covid-19 chest x-ray images classification: Using smote and weighted loss,” Applied Soft Computing, vol. 129, p. 109588, 2022.

\bibitem{ref54}
Y. Wu and H. Tan, “Short-term traffic flow forecasting with spatial-temporal correlation in a hybrid deep learning framework,” arXiv preprint arXiv:1612.01022, 2016.

\bibitem{ref55}
M. Li and Z. Zhu, “Spatial-temporal fusion graph neural networks for traffic flow forecasting,” in Proceedings of the AAAI conference on artificial intelligence, vol. 35, no. 5, 2021,pp. 4189–4196.









\bibitem{refBSWC2022}
J. Bi, Q. Sai, F. Wang, Y. Chen et al., “Identification of working trucks and critical path nodes for construction waste transportation based on electric waybills: A case study of shenzhen, china,” Journal of Advanced Transportation, vol. 2022, 2022.

\bibitem{refSDFK2018}
A. Sulemana, E. A. Donkor, E. K. Forkuo, S. Oduro-Kwarteng et al., “Optimal routing of solid waste collection trucks: A review of methods,” Journal of Engineering, vol. 2018, 2018.
\bibitem{refPTG2016}
R. Pérez-Torres, C. Torres-Huitzil, and H. Galeana-Zapién, “Full on-device stay points detection in smartphones for location-based mobile applications,” Sensors, vol. 16, no. 10, p. 1693, 2016.













\end{thebibliography}
\end{document}